%% file: main.tex
\documentclass[sigconf]{acmart}

\AtBeginDocument{%
  \providecommand\BibTeX{{%
    \normalfont B\kern-0.5em{\scshape i\kern-0.25em b}\kern-0.8em\TeX}}}

\setcopyright{acmcopyright}
\copyrightyear{2022}
\acmYear{2022}
\acmDOI{10.1145/3503161.3548211}
\acmConference[MM '22]{Proceedings of the 30th ACM International Conference on Multimedia}{October 10--14, 2022}{Lisbon, Portugal}
\acmBooktitle{Proceedings of the 30th ACM International Conference on Multimedia (MM '22), October 10--14, 2022, Lisbon, Portugal}
\acmPrice{15.00}
\acmISBN{978-1-4503-9203-7/22/10}

\usepackage{subfigure,epsfig}
\usepackage{algorithm}
\usepackage{algorithmic}
\usepackage{multirow}
\usepackage{float}
\usepackage{url}
\usepackage{footnote}

\usepackage{mathrsfs}

\usepackage{amsthm,amsmath}
\usepackage{mathrsfs}

\usepackage{enumitem}

\usepackage{array}
\usepackage{booktabs}

\setlist[itemize]{leftmargin=*}
\setlist[enumerate]{leftmargin=*}

\newcommand{\ie}{\emph{i.e., }}
\newcommand{\eg}{\emph{e.g., }}
\newcommand{\etal}{\emph{et al.}}

\def\BibTeX{{\rm B\kern-.05em{\sc i\kern-.025em b}\kern-.08emT\kern-.1667em\lower.7ex\hbox{E}\kern-.125emX}}

\begin{document}
\fancyhead{}
\title{Counterfactual Reasoning for Out-of-distribution\\ Multimodal Sentiment Analysis}

\author{Teng Sun}
\affiliation{%
	\institution{Shandong University}
 \country{}
}\email{stbestforever@gmail.com}

\author{Wenjie Wang$^*$}
\affiliation{%
	\institution{National University of Singapore}
	 \city{}
  \country{}
}\email{wenjiewang96@gmail.com}

\author{Liqiang Jing}
\affiliation{%
	\institution{Shandong University}
	 \city{}
  \country{}
}\email{jingliqaing6@gmail.com}

\author{Yiran Cui}
\affiliation{%
	\institution{Shandong University}
	 \city{}
  \country{}
}\email{conyrol120@gmail.com}

\author{Xuemeng Song}
\affiliation{%
	\institution{Shandong University}
	 \city{}
  \country{}
}\email{sxmustc@gmail.com}

\author{Liqiang Nie$^*$}
\affiliation{%
	\institution{\mbox{Harbin Institute of Technology (Shenzhen)}}
	\city{}
  \country{}
}\email{nieliqiang@gmail.com}
\def\authors{Teng Sun, Wenjie Wang, Liqiang Jing, Yiran Cui, Xuemeng Song, and Liqiang Nie}
\renewcommand{\shortauthors}{Teng Sun et al.}

\thanks{*Wenjie Wang (wenjiewang96@gmail.com) and Liqiang Nie (nieliqiang@gmail.com) are corresponding authors.}

\begin{abstract}
Existing studies on multimodal sentiment analysis heavily rely on textual modality and unavoidably induce the spurious correlations between textual words and sentiment labels. This greatly hinders the model generalization ability. To address this problem, we define the task of out-of-distribution (OOD) multimodal sentiment analysis. This task aims to estimate and mitigate the bad effect of textual modality for strong OOD generalization. To this end, we embrace causal inference, which inspects the causal relationships via a causal graph. From the graph, we find that the spurious correlations are attributed to the direct effect of textual modality on the model prediction while the indirect one is more reliable by considering multimodal semantics. Inspired by this, we devise a model-agnostic counterfactual framework for multimodal sentiment analysis, which captures the direct effect of textual modality via an extra text model and estimates the indirect one by a multimodal model. During the inference, we first estimate the direct effect by the counterfactual inference, and then subtract it from the total effect of all modalities to obtain the indirect effect for reliable prediction. Extensive experiments show the superior effectiveness and generalization ability of our proposed framework.

\end{abstract}


\vspace{-5pt}


\maketitle

\input{1_introduction.tex}
\input{2_related.tex}

\input{3_method.tex}

\input{4_experiment.tex}
\input{5_conclusion.tex}

\begin{acks}
This work is supported by the National Natural Science Foundation of China, No.:U1936203;  Beijing Academy of Artificial Intelligence (BAAI).
\end{acks}

\bibliographystyle{ACM-Reference-Format}
\balance
\bibliography{reference}
\end{document}

%% file: 1_introduction.tex
\section{Introduction}

Sentiment analysis, as a typical language understanding task, aims to automatically recognize the subjective opinions from users' reviews (\eg positive, neutral, or negative attitudes). 
Due to its wide applications in social media~\cite{DBLP:journals/tomccap/SunWSFN22} and recommender systems~\cite{DBLP:conf/sigir/ZhengGCYJ20}, sentiment analysis has attracted extensive attention from both academia and industry.
Early work mainly relies on the textual information for sentiment analysis~\cite{bert}. 
However, with the boom of mobile social media, more users tend to present their opinions in more diverse modalities, like images, audios, and videos.
For example, facial expressions and the tone of voice in videos usually express emotional tendencies, which alleviate the ambiguity of textual modality.
Inspired by this, the research focus has been recently shifted from the textual sentiment analysis to the multimodal one~\cite{DBLP:journals/corr/ZadehZPM16}.

Existing research on Multimodal Sentiment Analysis (MSA) mainly aims to mitigate the modality gap and learn discriminative multimodal representations. According to the fusion manner, they generally fall into two categories: early-fusion and late-fusion methods. 
The former integrates multimodal features in the shallow layers of neural networks before reasoning.
For example, Tsai \etal~\cite{mult} 
fused the raw features of different modalities with cross-modal attentive mechanisms, and then recognized the sentiments from the fused embeddings. 
By contrast, the latter learns the intra-modal representation separately, and performs the cross-modal analysis independently. For example, Liu \etal~\cite{lmf} first extracted the unimodal representation from each audio, video and text by three sub-networks without cross-modal interactions, and then leveraged a low-rank model to fuse the sentiment analysis results yielded by each modality.

\begin{figure}
\setlength{\abovecaptionskip}{0cm}
\setlength{\belowcaptionskip}{-0.4cm}
    \centering
    \includegraphics[scale=0.55]{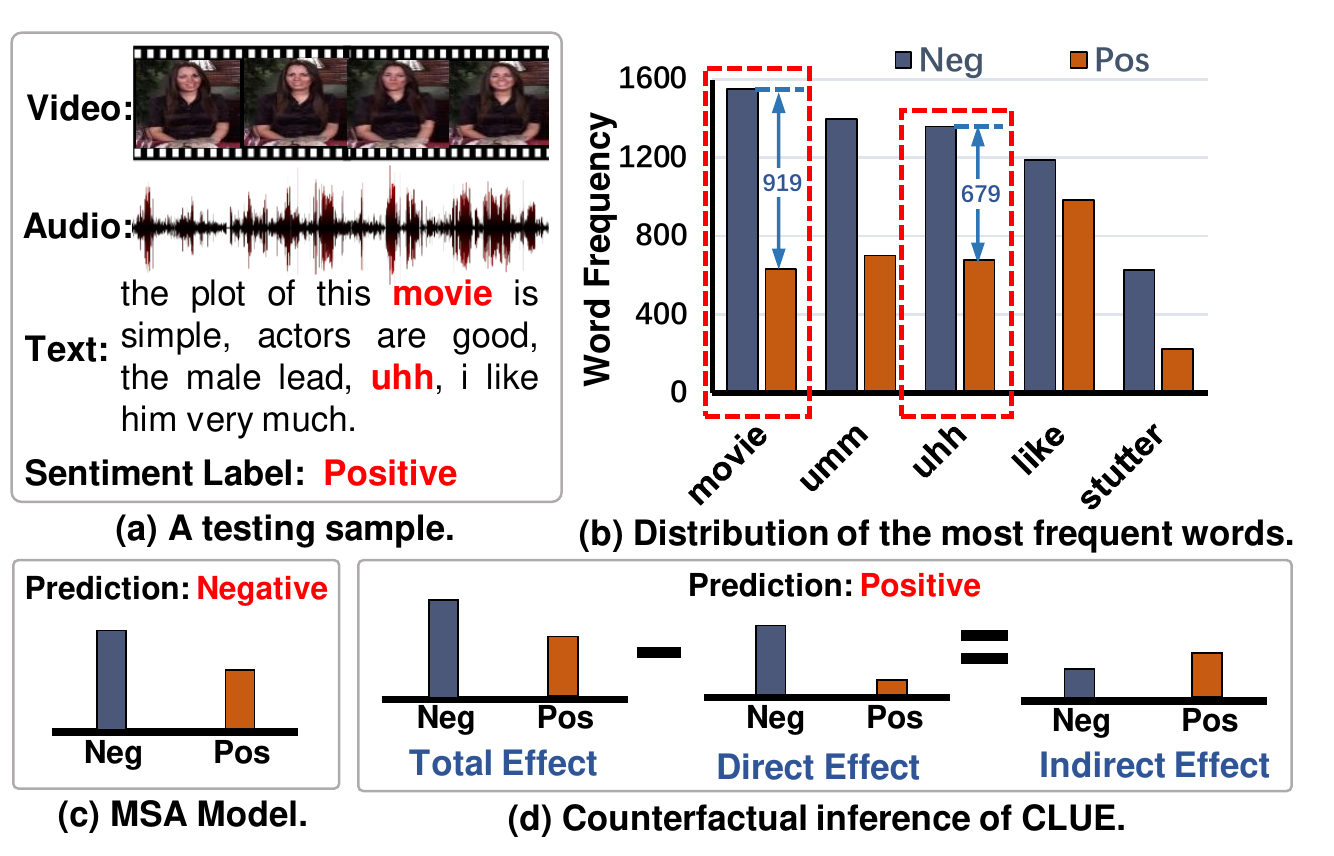}
    \caption{A testing sample, the distribution of the top-5 most frequent words in MOSEI~\cite{DBLP:conf/acl/MorencyCPLZ18}, and the predictions of MSA model and CLUE for the testing sample.}
    \label{fig:intro}
\end{figure}

Despite their compelling progress, existing studies usually suffer from fitting the spurious correlations in textual modality~\cite{DBLP:conf/mm/HazarikaZP20}.
As shown in Figure~\ref{fig:intro}(b), some words in textual modality seem strongly correlated with the sentiment label, which however are likely to be unreliable. 
For example, in the training set, the word ``movie'' appears in negative samples more frequently than that in the positive ones, making the model more likely to classify the testing samples with ``movie'' as negative. 
Indeed, ``movie'' is not a reliable cue for identifying negative samples. Such correlations are known as spurious correlations~\cite{causality} in causality, showing the bad effect of the textual modality. 
Deep neural networks are easy to adapt such correlations for prediction due to the short-cut bias~\cite{DBLP:journals/natmi/GeirhosJMZBBW20}, degrading the generalization ability in the out-of-distribution (OOD) scenarios where the correlations between textual words and labels differ from those in the training set.
To measure the true generalization ability of the model, we propose a novel OOD MSA task to evaluate whether MSA models can reduce the bad effect of textual modality in the OOD scenarios.

To alleviate the bad effect, an intuitive idea is to collect an unbiased dataset without spurious correlations. However, it is infeasible because various correlations between extensive words and the sentiment labels commonly exist in the real world.
Worse still, we cannot abandon the textual modality for sentiment analysis, considering its rich sentimental cues. 
In this light, our objective turns to eliminate the bad effect caused by spurious correlations while keeping the valuable semantics in textual modality. 
The keys to achieve this objective are threefold: 1) disentangling the good and bad effects of textual modality on the model prediction, 2) mitigating the bad effect for stronger OOD generalization, and 3) utilizing multimodal cues to alleviate the textual correlations. 

To this end, we resort to causal inference~\cite{glymour2016causal}. In particular, we utilize a causal graph to inspect the causal relationships between multiple modalities and the model prediction. From the graph, we find that multimodal cues are the key to disentangle the good and bad effects.
As described in Section~\ref{sec:clue}, we discover two kinds of effects of textual modality on the model prediction: direct and indirect. 
The former implies a shortcut from textual words to the prediction, exhibiting the spurious correlations in textual modality (\ie the bad effect).
As for the latter, it extracts reliable textual semantics by considering the multimodal cues. For example, the excited tone and the smiling face in the video well guide the model to focus on the word ``great'' and filter out the spurious correlations caused by
``movie''. As such, the indirect effect learned from the cross-modal alignment represents the good effect of texts.
Traditional MSA models are blindly optimized, resulting in correlation fitting between multiple modalities and sentiment labels, whereby two effects are mixed and the optimization hence inevitably favors the spurious correlations in the shortcut~\cite{DBLP:journals/natmi/GeirhosJMZBBW20}.

To disentangle the two effects and mitigate the bad one, we devise a model-agnostic CounterfactuaL mUltimodel sEntiment (CLUE) framework.
Specifically, CLUE calculates the indirect effect of textual modality by considering cross-modal modeling with a MSA model. Meanwhile, CLUE estimates the direct effect by adding an extra text model during training. 
In the inference stage, CLUE estimates the direct effect by counterfactual inference, which imagines \textit{what the prediction would be if the model had only seen the effect of texts}. 
Thereafter, CLUE mitigates such a direct effect and utilizes the indirect one of multiple modalities to alleviate the influence of spurious correlations in texts. 
To evaluate the generalization ability, we construct the OOD testing set
with significant correlation shifts.
Besides, we instantiate our CLUE framework on several latest MSA models. The empirical results on two benchmarks show that CLUE is able to significantly strengthen the generalization ability of the MSA models on the OOD testing set while maintaining competitive performance on the biased one.

To sum up, our contributions are threefold.
\begin{itemize}
    \item We define a novel OOD MSA task, which points out the spurious correlations in textual modality and highlights the necessity of strong OOD generalization abilities.
    \item We devise a model-agnostic CLUE framework. It strengthens the existing MSA models via capturing the causal relationships in the training set and mitigating the bad effect of textual modality by the counterfactual inference.
    \item We conduct extensive experiments on two benchmark datasets,
    and the results demonstrate the superior effectiveness and generalization ability of CLUE. As a byproduct, we have released the source code to facilitate the reproduction\footnote{\url{https://github.com/Teng-Sun/CLUE_model}.}.
\end{itemize}

%% file: 2_related.tex
\section{Related Work}

\vspace{5pt}
$\bullet$ \noindent\textbf{Multimodal Sentiment Analysis.}
The goal of MSA is to regress or classify the overall sentiment of an utterance via acoustic, visual, and textual cues.
Over the past decades, statistical knowledge and shallow machine learning algorithms have been frequently utilized in MSA~\cite{RF,bayes}. As shallow machine learning is limited by feature engineering and massive manual work for data annotation, 
deep learning becomes the mainstream technique for MSA~\cite{DBLP:conf/mm/ZhangLZZ19,GAN,trans,DBLP:conf/mm/0001F21} in an end-to-end manner~\cite{DBLP:conf/mm/YuanLXY21}. 


In a sense, the key to developing deep learning-based MSA models lies in two aspects: multimodal representation learning and cross-modal fusion. Studies on the former are in three variants:
~1)~Translation-based models translate one modality to another\cite{seq2seq,cyclic,autoencoder}.
~2)~Correlation-based models learn cross-modal correlations by using Canonical Correlation Analysis~\cite{correlation}. 
And~3)~the shared subspace learning models map all the modalities simultaneously into a new shared subspace via specific algorithms, like adversarial learning~\cite{DBLP:conf/mm/NiSZZG21}. 
Another line of MSA focuses on developing sophisticated fusion techniques. 
So far, the existing fusion strategies can be categorized into feature-level and decision-level fusion. The former focus on fusing the representations of different modalities with the help of methods such as the tensor fusion network~\cite{tfn}.
 Regarding the decision-level fusion, the features of different modalities are classified independently based on their individual features, and then all classification results are integrated into a final prediction~\cite{latefusion}. 

Although existing studies have achieved great success, they ignore the spurious correlations between texts and the sentiment labels. As such, we proposed a new task, named OOD MSA, with a counterfactual framework to improve the OOD generalization ability of MSA models.

\vspace{5pt}
$\bullet$ \noindent\textbf{Causal Inference.}
Causal inference has been widely-used across many tasks to improve the robustness of deep learning models, including visual question answering~\cite{niu2021counterfactual, li2022invariant,li2021interventional}, visual commonsense reasoning~\cite{zhang2021multi},
~recommended system~\cite{wang2022causal}, 
~and text classification~\cite{qian2021counterfactual}.
Regarding the theory of causal inference, counterfactual reasoning enables models to 
achieve counterfactual imagination and estimate the causal effect.
The methods of counterfactual reasoning can be broadly categorized into three strategies: 
$1)$~Some efforts estimated the path-specific causal effect by counterfactual inference such as natural direct effect and total indirect effect~\cite{niu2021counterfactual}, and then mitigate some effects to achieve unbiased prediction~\cite{DBLP:conf/sigir/WangF0ZC21}. $2)$~Some researchers trained a ``poisonous'' model regardless of the dataset biases and utilized $do$ calculus to make unbiased predictions in the test phase~\cite{qian2021counterfactual}.
~$3)$~Existing studies also consider generating counterfactual samples for a factual sample and compare the counterfactual and factual samples to retrospect the original model prediction~\cite{lanunder}.

As for the MSA task, the model is at risk of capturing the bias from the textual modality and may make predictions based on the spurious correlations between texts and labels. 
~Therefore, different from existing work, we attempted to apply counterfactual reasoning to MSA for mitigating the spurious correlations. 

%% file: 3_method.tex
\begin{figure}
\setlength{\abovecaptionskip}{-0.10cm}
\setlength{\belowcaptionskip}{-0.30cm}
    \centering
    \includegraphics[scale=0.55]{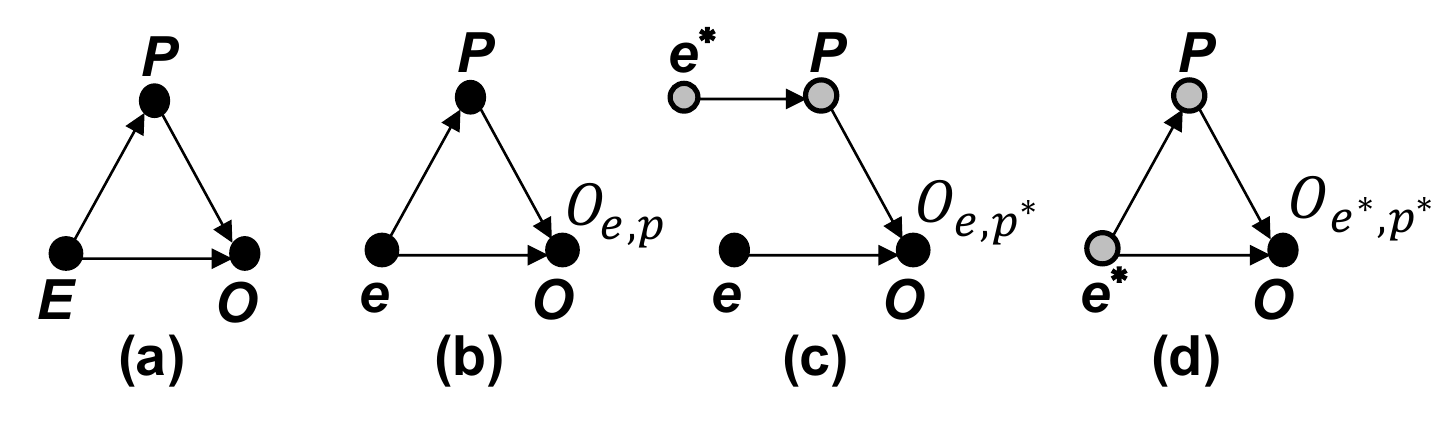}
    \caption{(a) The causal graph of the admission outcome, where the admission outcome of graduate ($O$) is directly affected by the experience ($E$) and publication ($P$). Counterfactual notations $O_{e, p}$, $O_{e, p^*}$ and $O_{e^*, p^*}$ are illustrated in (b), (c) and (d), respectively. }
    \label{fig:preliminary}
\end{figure}

\section{Preliminary} \label{sec:preliminary}
In this section, we refer to this work~\cite{niu2021counterfactual} to detail the key concepts and symbols related to causal inference.

\textbf{Casual graph} is a directed acyclic graph $\mathcal{G}=\{\mathcal{N}, \mathcal{E}\}$ which reveals the causal relationships among variables. 
Meanwhile, the nodes $\mathcal{N}$ denote the variables and the edges $\mathcal{E}$ represent the causal relationships among variables. 
An example of a causal graph about the admission outcome 
~is illustrated in Figure~\ref{fig:preliminary}(a). The casual graph mainly reflects the relationships among the experience ($E$), the publication ($P$), and the admission outcome ($O$): 1) $O$ is directly affected by $E$ and 2) indirectly influenced by $E$ via the mediator $P$. 

\textbf{Counterfactual Inference.}
The counterfactual inference is to answer the counterfactual question based on factual observations, \ie \textit{``what the outcome would be if the treatment was different from the factual one?''}~\cite{causality}. The treatment can be seen as assigning a specific value to the random variable.
We utilize the lowercase letters to denote the specific values of random variables. 
Furthermore, we introduce structural equations~\cite{causality} to formulate the causal relationships among causal variables for the convenience of subsequent understanding.  
For example, the values $P$ and $O$ under the conditions of $E=e$ and $P=p$ can be formulated as follows,
\begin{equation}
P_e = p = f_P (E=e), O_{e,p} = f_O (E=e, P=p), 
\label{eq:notation}
\end{equation}
where $P_{e}$ denotes what the publication of someone would be if he/she has the experience $e$ 
~, and $O_{e,p}$ indicates what the admission outcome of someone would be if he/she has the experience $e$ 
~and publication $p$ 
~(see Figure~\ref{fig:preliminary}(b)). $f_P(\cdot)$ and $f_O(\cdot)$ denote the structural equations for the variables $P$ and $O$, respectively, which can be learned from the observational data~\cite{causality}. 

Then, we can describe the counterfactual and factual situations by notations in the structural equations.
For example, Figure~\ref{fig:preliminary}(c) illustrates ``what the admission outcome would be if someone only has the publication without the experience?'', where $O$ receives $E=e$ by $E \rightarrow O$ , while $P$ receives $E=e^*$ through $E\rightarrow P$. $E=e^*$ denotes no experience whatsoever, otherwise, $E=e$.

\textbf{Causal effect.}
Causal effect refers to the difference in the outcome variable when the reference value of the variable change to a specific one~\cite{1986A}, it is also named as Total Effect~(TE). 
For example, TE of the experience $E=e$ (\ie the treatment variable) on the admission outcome $O$ can be formulated as follows,
\begin{equation}
\small
 \left\{
 \begin{aligned}
     \text{TE} &= O_{e,p} - O_{e^*,p^*} =  f_O (E=e, P=p) - f_O (E=e^*, P=p^*),\\
     p^* &= P_{e^*}= f_P (E=e^*),
\end{aligned}
 \right.
\label{eq:pre_te}
\end{equation}
where $O_{e^*,p^*}$ denotes the reference situation, $E$ and $P$ are set as the reference values $e^*$~(\ie no experience) and $p^*$ (\ie the publication without experience), respectively. 
The corresponding illustration in the causal graph is presented in Figure~\ref{fig:preliminary}(d). 
Specifically, $O_{e^*, p^*}$ denotes what the admission outcome of someone would be if he/she has no experience (\ie $E=e^*$). 
TE (\ie Eqn. ($\ref{eq:pre_te}$)) measures the change of $O$ when $E$ changes from having no experience (\ie $E=e^*$) to having it (\ie $E=e$). 
Meanwhile, the Natural Direct Effect (NDE) and Total Indirect Effect (TIE) together make up TE~\cite{causality} as follows,
\begin{equation}
    \text{TE} = \text{NDE} + \text{TIE}. \label{eq:infer}
\end{equation}
Specifically, NDE denotes the effect of $E=e$ on $O$ along the direct path $E\rightarrow O$ without considering the indirect effect along $E\rightarrow P \rightarrow O$. TIE measures the difference between TE and NDE, \ie the change of $O$ when $E$ changes from having no experience (\ie $E=e^*$) to having it (\ie $E=e$) along the indirect effect path $E\rightarrow P \rightarrow O$. 
For example, NDE of $E=e$ on $O$ can be calculated by:
\begin{equation}
    \text{NDE} = O_{e,p^*} - O_{e^*,p^*}, 
\label{eq:pre_nde}
\end{equation}
where $O_{e,p^*}$ denotes a counterfactual situation, $E$ is set to $e$, and $P$ is set to $p^*$, \ie the value of $P$ under $E=e^*$ (see Figure~\ref{fig:preliminary}(c)).
Based on Eqn. ($\ref{eq:infer}$) and ($\ref{eq:pre_nde}$), we can obtain TIE of $E=e$ on $O$ by subtracting NDE from TE:
\begin{equation}
    \text{TIE} = \text{TE} - \text{NDE} = O_{e,p} - O_{e,p^*}.
\end{equation}


\begin{figure}
\setlength{\abovecaptionskip}{-0.10cm}
\setlength{\belowcaptionskip}{-0.3cm}
    \centering
    \includegraphics[scale=0.65]{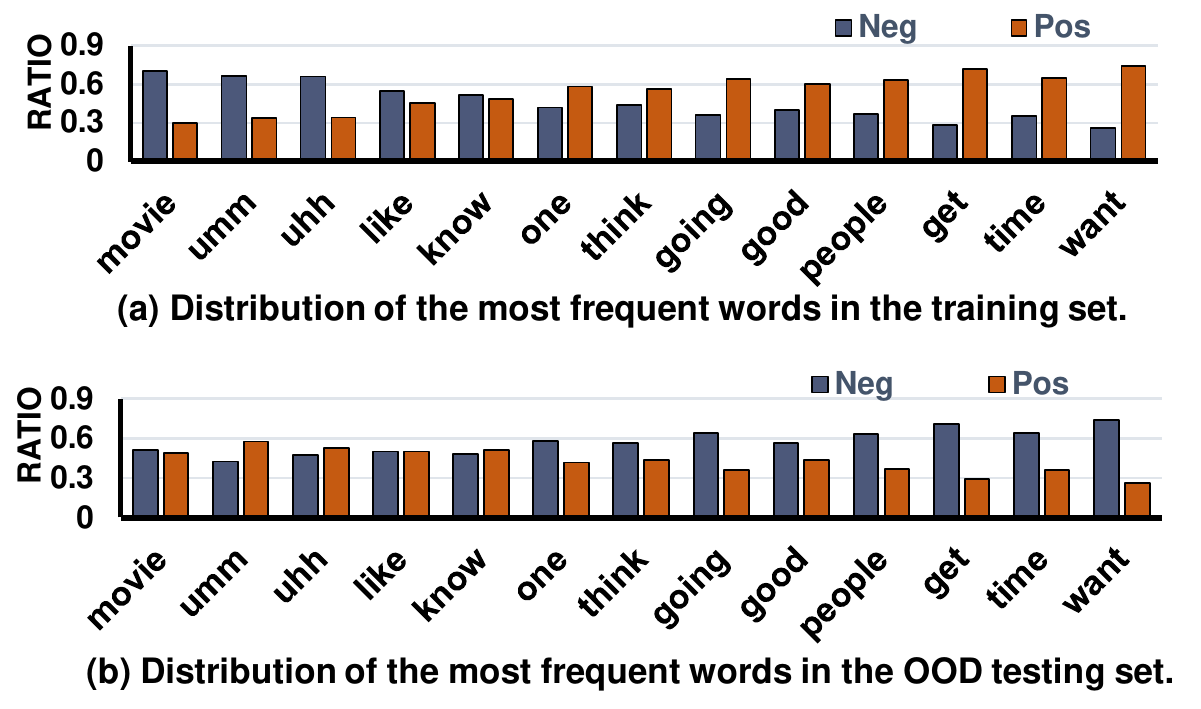}
    \caption{Distribution of the most frequent words in the training set and the OOD testing set of MOSEI. The y-axis shows the proportion of samples with the target word (\eg ``movie'') in the positive or negative categories.}
    \label{fig:ood}
\end{figure}

\section{Methodology}

We first formulate the task of OOD MSA, and then detail the proposed CLUE framework.

\subsection{Task Formulation}

\textbf{Traditional MSA Task.} A dataset with $N$ training samples can be formulated as $\mathcal{D}=\{(t_1, a_1, v_1, y_1), \cdots, (t_N, a_N, v_N, y_N)\}$. Each quadruple $(t_i, a_i, v_i, y_i)$ is associated with a video segment, whereby $t_i$, $a_i$, $v_i$, and $y_i$ denotes the text, audio, video, and sentiment label of the $i$-th sample, respectively. 
The MSA aims to devise a multimodal model $\mathcal{F}_\theta$, which jointly considers the three modalities (\ie $t_i, a_i,$ and $v_i$) as inputs and predicts the sentiment label (\ie $y_i$) as follows,
\begin{equation}\label{eqn:MSA}
    \hat{y}_i = \mathcal{F}_{\theta} (t_i, a_i, v_i),
\end{equation}
where $\hat{y}_i$ denotes the predicted probability of sentiment labels for $i$-th sample and $\theta$ represents the learnable parameters.





\vspace{3pt}
\textbf{OOD MSA Task.}
Traditional MSA models tend to heavily rely on textual modality for the sentiment analysis~\cite{DBLP:conf/mm/HazarikaZP20} because salient textual semantics (\eg ``it is great.'') easily reflect the sentiment tendency. 
However, there exist some spurious correlations between textual words and sentiment labels due to the selection bias~\cite{causality} when collecting the dataset $\mathcal{D}$. For example, the word ``movie'' is highly correlated with the negative label in the MOSEI dataset as shown in Figure~\ref{fig:ood}(a). 
The model trained with such biased data may prefer to classify the samples with ``movie'' as negative, which deteriorates the generalization performance of MSA models.

To reduce the influence of spurious correlations, we define an OOD MSA task, which aims to evaluate the OOD generalization performance of different MSA methods on the OOD testing set. 
Specifically, we develop an algorithm (illustrated in Section \ref{sec:ood_testing}) to automatically construct the OOD testing set for each biased dataset, with significantly different word-sentiment correlations from the training one. 
The different distributions between the training and OOD testing sets can validate whether the MSA models are able to mitigate the effect of spurious correlations effectively. 

\begin{figure}
\setlength{\abovecaptionskip}{0cm}
\setlength{\belowcaptionskip}{-0.30cm}
\centering
\includegraphics[scale=0.6]{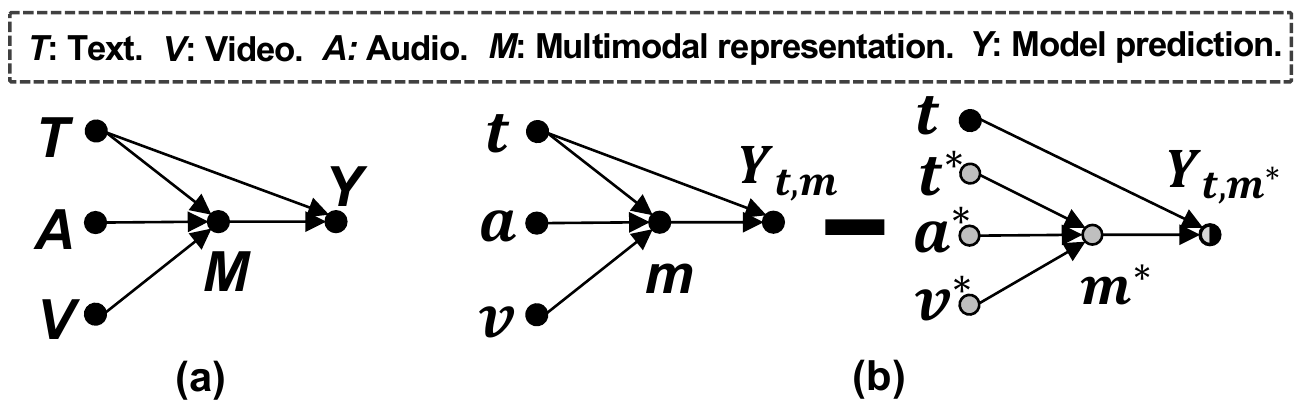}
\caption{ (a) The causal graph in the MSA. And (b) the illustration of counterfactual inference.}
\label{fig:causal_msa}
\end{figure}

\begin{figure*}
\setlength{\abovecaptionskip}{0cm}
\setlength{\belowcaptionskip}{-0.30cm}
    \centering
    \includegraphics[scale=0.60]{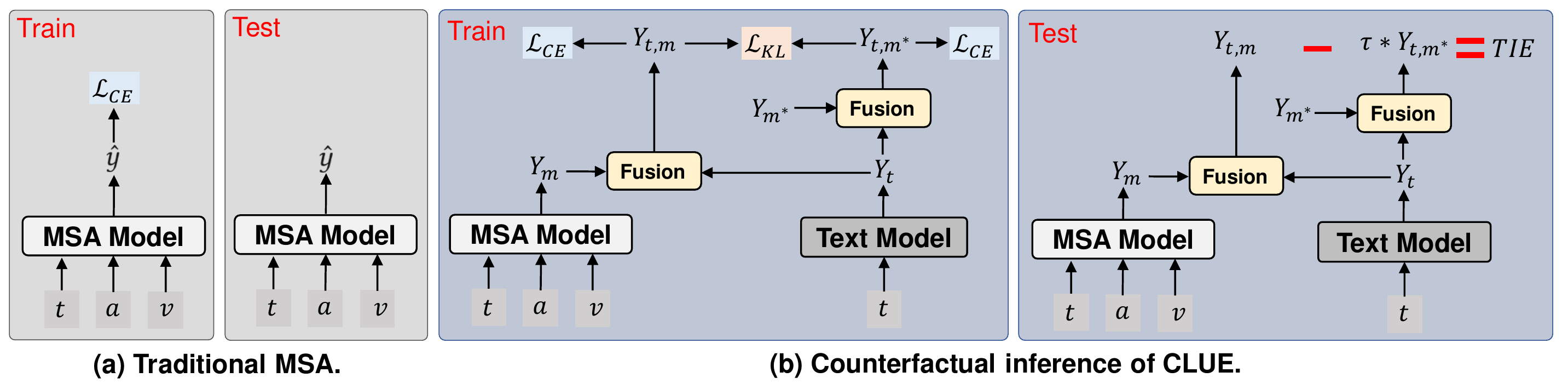}
    \caption{Illustration of traditional MSA models and the counterfactual inference of CLUE.}
    \label{fig:framework}
\end{figure*}

\subsection{CLUE Framework} \label{sec:clue}
We first inspect the causal relationships in the MSA task, and then detail the CLUE framework. 
\subsubsection{\textbf{Causal Graph in the MSA Task}} \label{sec:causal_msa}
As shown in Figure~\ref{fig:causal_msa}(a), we utilize a causal graph to describe the causal relationships in the MSA task, \ie the relationships among the text $T$, audio $A$, video $V$, multimodal representation $M$, and model prediction $Y$. 
The explanations for the causal relationships are as follows:
\begin{itemize}
    \item $T \rightarrow Y$ is a shortcut between the text and the model prediction because textual modality provides salient features with spurious correlations for the model prediction. This shortcut describes the bad effect of textual modality due to the spurious correlations between texts and labels. 
    \item $(T, A, V) \rightarrow M \rightarrow Y$ denotes the indirect effect of $T$, $A$, and $V$ on the model prediction $Y$ via a mediator $M$. The multimodal representation $M$ is obtained by the cross-modal modeling, which extracts more reliable textual semantics based on the alignment among different modalities. For example, the excited tone in the voice and the smiling face in the video guide $M$ to focus on the word ``great'' in text and filter out the spurious correlations caused by ``movie''. Therefore, the indirect path $T\rightarrow M\rightarrow Y$ presents a good effect of texts. 
\end{itemize}

\textbf{Causal View of Traditional MSA Models.}
Traditional MSA models ignore the causal relationships in Figure \ref{fig:causal_msa}, and thus blindly fit the correlations between the features of three modalities and sentiment labels as shown in Eqn. (\ref{eqn:MSA}). As such, they mix the direct and indirect effects of textual modality on the model prediction, and then inevitably utilize the spurious correlations in the direct effect for unreliable predictions.

\subsubsection{\textbf{Counterfactual MSA Framework.}}

Inspired by previous work~\cite{niu2021counterfactual,DBLP:conf/sigir/WangF0ZC21} embracing causal inference, we propose to model the causal relationships in Figure \ref{fig:causal_msa} based on the CLUE framework, which captures both the direct and indirect effects of textual modality. 
Following the notations in Section \ref{sec:preliminary}, the abstract format of causal relationships is:
\begin{equation}
 \left\{
 \begin{aligned}
&Y_{t, m} = f_Y (T=t, M=m), \\\
&m = f_M (T=t, A=a, V=v).
\end{aligned}\label{eq:notation_msa}
 \right.
\end{equation}
where $Y_{t,m}$ denotes the model prediction under the situation of $M=m$ and $T=t$.
Although Eqn. (\ref{eq:notation_msa}) provides the accurate formulation of causal relationships, $Y_{t,m}$ still suffers from the bad effect (\ie the direct effect) of $T=t$. This is because $Y_{t,m}$ contains both the direct and indirect effects. 
To disentangle them, we estimate NDE and TIE of textual modality on the model prediction separately. To accomplish this, we first calculate TE and NDE of texts, and then subtract NDE from TE to obtain TIE for the inference of CLUE. 

Following the definition of causal effects in Section~\ref{sec:preliminary}, TE on the model prediction $Y$ under the situation of $T=t$, $A=a$ and $V=v$ can be calculated by
\begin{equation}\small
\begin{aligned}
    \text{TE} &= Y_{t,m} - Y_{t^*, m^*} = f_Y(T=t, M=m) - f_Y(T=t^*, M=m^*),
\end{aligned}
\end{equation}
where $m^* = f_M (T = t^*, A = a^*, V = v^*)$. Meanwhile, the symbols $t^*$, $a^*$ and $v^*$ denote the reference values of $T$, $A$, and $V$, respectively. In this work, we define the reference values as the situations where the features are not given, \ie using a dummy vector as the features of $M$. 
Next, we estimate the NDE of textual modality. 
Formally, the NDE of $T=t$ on the model prediction $Y$ can be formulated as follows,
\begin{equation}
    Y_{t,m^*} - Y_{t^*, m^*} =f_Y(T=t, M=m^*) -f_Y(T=t^*, M=m^*),
\end{equation}
where $T$ is set to $t$ in the path $T\rightarrow Y$ while keeping $T=t^*$ along the path $T\rightarrow M \rightarrow Y$.  $Y_{t,m^*} = f_Y(T=t, M=m^*)$ denotes the corresponding model prediction $Y$. 
$Y_{t,m^*}$ answers a counterfactual problem: \textit{what the model prediction would be if the MSA model only had seen the direct effect of the textual modality}.
To mitigate the bad effect of $T=t$ in the direct path, in the inference stage, we subtract NDE from TE to calculate TIE of three modalities~(see Figure~\ref{fig:causal_msa}(b)):
\begin{equation}
\begin{aligned}
    \text{TIE} = \text{TE} - \text{NDE}=Y_{t,m} - Y_{t, m^*}, \label{eq:tie}
\end{aligned}
\end{equation}
which mitigates the bad effect (\ie NDE) caused by the spurious correlations in the shortcut $T\rightarrow Y$. As such, we utilize the reliable indirect effect (\ie TIE) for the unbiased prediction by counterfactual inference.

Figure~\ref{fig:intro}(d) illustrates the prediction result from a sample with the word ``movie''. In the prediction of NDE, this sample has a higher probability over the negative category due to the spurious correlation between words and the negative label. However, NDE is subtracted from TE, leading to a relatively higher probability over the positive category.




\subsection{Instantiation of CLUE}
To calculate TIE of textual modality, we implement the CLUE framework (see Figure~\ref{fig:framework}) based on the causal graph in Figure \ref{fig:causal_msa}(a). 

\textbf{Model Architecture}. We implement two structural equations $f_Y (T=t, M=m)$ and $f_M (T=t, A=a, V=v)$ according to Eqn.~($\ref{eq:notation_msa}$). For $f_M (T=t, A=a, V=v)$, we utilize the cross-modal feature extractors in conventional MSA models, which take the text, audio, and video as inputs to calculate $M$. As to $f_Y (T=t, M=m)$, we implement it in a late-fusion manner inspired by existing research~\cite{niu2021counterfactual}. 
This manner can well utilize the strong representation capability of existing MSA models~\cite{DBLP:conf/sigir/WangF0ZC21}. We thus reach the following,
\begin{equation}
    f_Y(T=t, M=m) = h(Y_t, Y_m), \label{fusion}
\end{equation}
where $Y_t = g_Y (T=t)$ and $Y_m = g_Y(M=m) = g_Y(T=t, A=a, V=v)$ are the predicted probabilities by two neural models with different inputs, $g_Y (T=t)$ can be implemented by any traditional unimodal model, \eg BERT~\cite{bert}, $g_Y(T=t, A=a, V=v)$ can be instantiated by any multimodal MSA model, \eg MISA~\cite{DBLP:conf/mm/HazarikaZP20}, and $h(\cdot)$ is a fusion function to combine the predictions $Y_t$ and $Y_m$. In this work, we adopt the widely-adopted SUM fusion strategy:
\begin{equation}
Y_{t,m} = f_Y(T=t, M=m) = \text{SUM}(Y_t, Y_m) = \log \sigma(Y_t + Y_m),
\label{eq:fusion}
\end{equation}
where $\sigma$ denotes the sigmoid function. In fact, since $Y_{t,m}$ represents the prediction probability and its sum should be one, we should apply the softmax operation after Eqn. (12).
Meanwhile, whether using the softmax operation will not affect the prediction result, because we select the class with the maximal prediction score as the sentiment prediction.

\textbf{Training}. Towards optimization, we adopt 
the cross-entropy ($CE(\cdot)$) loss for $Y_{t,m}$ and $Y_{t, m^*}$ as follows:
\begin{equation}
    \mathcal{L}_{CE} = \alpha *  \text{CE}(Y_{t, m}, y) + \beta * \text{CE}(Y_{t, m^*}, y), \label{eq:celoss}
\end{equation}
where $\alpha$ and $\beta$ are the weights to balance two loss functions, and $y$ is the sentiment label for $(t, a, v)$. 
In addition to using $Y_{t, m}$ to optimize TE, we also introduce a CE loss to facilitate the NDE prediction $Y_{t, m^*}$ in the counterfactual world. 
We calculate $Y_{t, m^*}=h(Y_t, Y_{m^*})$ by Eqn. ($\ref{eq:fusion}$), where $Y_{m^*}$ indicates the MSA model without $(t, a, v)$ as inputs. 
Theoretically, we should utilize randomly initialized vectors or all-zero vectors to denote $t^*$, $a^*$ and $v^*$, and then feed them into $f_M$ to get $m^*$. Because only $t^*$, $a^*$ and $v^*$ affect the prediction of $Y_{m^*}$ and $Y_{m^*}$ is not affected by other features, we directly use a randomly initialized vector to represent the final result $Y_{m^*}$ (i.e., the calculation result of $t^*$, $a^*$ and $v^*$), and do not need to instantiate $t^*$, $a^*$ and $v^*$ in the real implementation.
We thus use a trainable vector $Y_{m^*}$ as the output of the MSA model, \ie a dummy vector shared for all samples.  

An inappropriate $Y_{m^*}$ may make the scales of TE and NDE different so that TIE would be dominated by one of them~\cite{niu2021counterfactual}.
Therefore, we leverage Kullback-Leibler (KL) divergence to minimize the difference between $Y_{t, m^*}$ and $Y_{t, m}$ to estimate $Y_{m^*}$:
\begin{equation}
    \mathcal{L}_{KL} = \text{KL}(Y_{t, m^*}, Y_{t, m}).
\end{equation}
To summarize, the final training loss is as follows:
\begin{equation}
    \mathcal{L} =  \sum_{(t_i,a_i,v_i,y_i)\in \mathcal{D}} \mathcal{L}_{CE}+ \gamma * \mathcal{L}_{KL}, \label{eqn:allloss}
\end{equation}
where $\gamma$ is the parameter to control the effect of $\mathcal{L}_{KL}$. We optimize CLUE in an end-to-end manner.

\textbf{Inference}. To leverage TIE for inference, the final prediction of CLUE can be obtained by, 
\begin{equation}
\begin{aligned}
\text{TIE} &= Y_{t,m} - \tau * Y_{t, m^*} = h(Y_t, Y_m) - \tau * h(Y_t, Y_{m^*}), \label{eq:msa_infer}
\end{aligned}
\end{equation}
where $\tau$ is the hyperparameter to control the degree of subtracting correlations between the textual modality and sentiment labels.

%% file: 4_experiment.tex
\section{Experiments}
In this section, we conducted extensive experiments to answer the following research questions:
\begin{itemize}
    \item \textbf{RQ1}. Does CLUE improve the performance of MSA models on the OOD testing?
    \item \textbf{RQ2}. How does CLUE perform on the biased testing, \ie the Independent and Identically Distributed (IID) testing?
    \item \textbf{RQ3}. How does each component of CLUE affect the results?
\end{itemize}

\subsection{Experimental Settings}

\subsubsection{\textbf{Datasets.}}
For evaluation, we adopted two publicly accessible benchmark datasets on MSA.

\textbf{MOSI~\cite{DBLP:journals/corr/ZadehZPM16}} is 
a collection of 2,199 subjective utterance-video clips from 93 YouTube monologue videos.
Each clip is labeled with a continuous sentiment score ranging from $-3$ (strongly negative) to $3$ (strongly positive). 

\textbf{MOSEI~\cite{DBLP:conf/acl/MorencyCPLZ18}} is an expanded version of MOSI with a larger number of utterance-video clips from more speakers and topics. Statistically, it comprises $23,453$ annotated utterance-video clips, derived from $5,000$ videos with $1,000$ distinct speakers and $250$ different topics.


\begin{algorithm}[!t]
\caption{IID and OOD Set Construction.}
\label{alg:ood}
\begin{flushleft}
\hspace*{0.02in} {\textbf{Input:}  The whole dataset $\mathcal{D}$, the pre-defined distribution difference $\boldsymbol{\phi}_{\Delta}$, the number of iterations $n$, simulated annealing temperature $\tau$, and the temperature decay rate $\alpha$.
}
\\
\hspace*{0.02in} {\textbf{Output:} IID set $\mathcal{D}_{iid}$ and OOD set $\mathcal{D}_{ood}$} . 
\end{flushleft}
\begin{algorithmic}[1]
\STATE{Get an IID set $\mathcal{D}_{iid}$ and OOD set $\mathcal{D}_{ood}$ by random splitting $\mathcal{D}$.}
\STATE {Compute distributions $\boldsymbol{\phi}_{iid}$ and $\boldsymbol{\phi}_{ood}$ of all words over different sentiment categories in $\mathcal{D}_{iid}$ and $\mathcal{D}_{ood}$, respectively.}
\STATE {Set $V = \Vert abs(\boldsymbol{\phi}_{iid} - \boldsymbol{\phi}_{ood}) - abs(\boldsymbol{\phi}_{\Delta})\Vert_1$.}

\REPEAT{
    \REPEAT{
        \STATE {Randomly swap samples between $\mathcal{D}_{iid}$ and $\mathcal{D}_{ood}$ by perturbation strategies\footnotemark{} to 
        a new IID set 
        $\hat{\mathcal{D}}_{iid}$ and OOD set $\hat{\mathcal{D}}_{ood}$.}
        \STATE {Calculate $\boldsymbol{\hat{\phi}}_{iid}$ ( $\boldsymbol{\hat{\phi}}_{ood}$) with $\hat{\mathcal{D}}_{iid}$ ($\hat{\mathcal{D}}_{ood}$), respectively.}
        \STATE {Set $\hat{V} = \Vert abs(\boldsymbol{\hat{\phi}}_{iid} - \boldsymbol{\hat{\phi}}_{ood}) - abs(\boldsymbol{\phi}_{\Delta})\Vert_1$.}
        \STATE {Get a random number $R$ and $0 \le R < 1$. }
        \IF{$V \ge \hat{V}$}
        \STATE{Set $\mathcal{D}_{iid}, \mathcal{D}_{ood}, V = \hat{\mathcal{D}}_{iid}, \hat{\mathcal{D}}_{ood}, \hat{V}$.}
        \ELSIF{$exp{((\hat{V} - V) / \tau)} > R$}
        \STATE{Set $\mathcal{D}_{iid}, \mathcal{D}_{ood}, V = \hat{\mathcal{D}}_{iid}, \hat{\mathcal{D}}_{ood}, \hat{V}$.}
        \ENDIF
    }
    \UNTIL{Swapping times reach $n$.}
    \STATE {Set $\tau = \tau * \alpha$.}
}
\UNTIL{Iteration times reach $n$.}
\end{algorithmic}
\end{algorithm}
\footnotetext{Referring to the study~\cite{simulated}, we utilized three perturbation strategies.}

\subsubsection{\textbf{Evaluation Tasks and Metrics.}}

Towards comprehensive evaluation, we assessed the MSA models with two widely-adopted tasks: $7$-class classification and $2$-class classification. 
The $7$ sentiment classes in the former task range from strongly negative ($-3$) to strongly positive ($3$) and the accuracy (\textit{Acc-7}) was adopted as the evaluation metric.
Pertaining to the latter task, we further investigated two settings: \textit{negative/non-negative} and \textit{negative/positive}, as each has been once studied by previous work~\cite{DBLP:conf/aaai/ZadehLPVCM18,mult}. 
For the \textit{negative/non-negative} setting, we classified the sentiment of a given sample as the negative or non-negative class, while for the \textit{negative/positive} setting, our goal is to assign each sample with a negative or positive sentiment.
Towards the evaluation on the binary task, we used both accuracy (denoted as \textit{Acc-2}) and $F1$ score.

\subsubsection{\textbf{IID and OOD Settings.}} \label{sec:ood_testing}
For each task, we performed experiments in both IID and OOD settings. In the former setting, the testing set shares the IID distribution with the training one, which is similar to many research~\cite{wei2021contrastive,wei2019mmgcn,hu2019hierarchical,hu2021efficient,Liu2021IMP_GCN}; Whereas in the latter, we expected that the sample distribution of each word over different sentiment categories in the testing set is as different as possible from the training set. In a sense, for each task, it requires four sets of samples from each dataset: IID training, IID validation, IID testing, and OOD testing sets. 
Similar to previous IID studies, we still adopted the output of the MSA model to search the optimal parameters with the IID validation set. 
In the testing phase, we used TE of all modalities $Y_{t,m}$ without mitigating the bias for the biased IID testing, while the counterfactual inference results (\ie TIE in Eqn. ($\ref{eq:msa_infer}$)) for the OOD testing.

\begin{table*}[t]
\setlength{\abovecaptionskip}{0cm}
\setlength{\belowcaptionskip}{0cm}
\centering
\caption{OOD testing performance (\%) comparison among different methods on MOSEI and MOSI datasets. For \textit{Acc-2} and \textit{F1}, ``$*$'' is calculated as ``negative/non-negative'' and ``$\S$'' is calculated as ``negative/positive''. The best result of each pair of the original MSA model and the model with CLUE is highlighted in bold. }
\resizebox{0.98\textwidth}{!}{
\begin{tabular}{c|ll|ll|l|ll|ll|l}
\toprule 
\multicolumn{1}{c|}{\multirow{3}{*}{Model}} & \multicolumn{5}{c|}{MOSEI} & \multicolumn{5}{c}{MOSI}  \\ 
& \multicolumn{4}{c}{2-class} & \multicolumn{1}{c|}{7-class} & \multicolumn{4}{c}{2-class} & \multicolumn{1}{c}{7-class}  \\ \cline{2-11} 
& \textit{Acc-2}$^*$ & \textit{F1}$^*$ & \textit{Acc-2}$^\S$ & \textit{F1}$^\S$  & \textit{Acc-7}& \textit{Acc-2}$^*$ & \textit{F1}$^*$ & \textit{Acc-2}$^\S$ & \textit{F1}$^\S$ &  \textit{Acc-7}  \\ \hline
TFN	 &71.23& 70.46&69.76  &69.02  &41.05  &73.02&72.93 &74.62 & 74.56  &32.95 \\ 
LMF	 &68.16&68.31&69.58 &69.58	&31.11	&73.54&73.40&75.27 &75.18 &29.10 \\ 
MulT 	&72.56& 72.44&73.73	&73.58&	40.58		&75.00&74.75 & 76.72	& 76.52		&29.80	\\ \hline
MAG-BERT	&74.59& 74.48&76.41	&76.27	&45.88		&75.57& 75.52&77.28	&77.26	&39.85 \\
$\quad$+CLUE (Ours)	&\textbf{78.34}$^{+3.75}$& \textbf{78.23}$^{+3.75}$&\textbf{80.51}$^{+4.10}$	&\textbf{80.46}$^{+4.19}$	&\textbf{48.66}$^{+2.78}$		&\textbf{77.25}$^{+1.68}$ & \textbf{77.46}$^{+1.94}$ &\textbf{78.65}$^{+1.37}$	&\textbf{78.83}$^{+1.57}$	&\textbf{40.75}$^{+0.90}$ \\ \hline 
MISA	&74.48&	74.39 &76.45&76.33&	43.15&		75.90& 75.82&77.39&	77.35&	38.05\\ 
$\quad$+CLUE (Ours)	&\textbf{77.17}$^{+2.69}$& \textbf{77.08}$^{+2.69}$& \textbf{78.77}$^{+2.32}$	&\textbf{78.74}$^{+2.41}$	&\textbf{46.86}$^{+3.71}$	&\textbf{78.25}$^{+2.35}$&\textbf{78.28}$^{+2.46}$ &\textbf{79.17}$^{+1.78}$	&\textbf{79.19}$^{+1.84}$ 	&\textbf{42.25}$^{+4.20}$ \\ \hline
Self-MM 	&74.68& 74.33&74.50	&74.22	&45.81 	&76.70& 76.68&78.12	&78.13		&40.25 \\ 
$\quad$+CLUE (Ours) 	&\textbf{77.76}$^{+3.08}$&\textbf{77.72}$^{+3.39}$ &\textbf{79.48}$^{+4.98}$	&\textbf{79.47}$^{+5.25}$	&\textbf{48.09}$^{+2.28}$	&\textbf{78.75}$^{+2.05}$& \textbf{78.75}$^{+2.07}$&\textbf{79.94}$^{+1.82}$	&\textbf{79.93}$^{+1.80}$	&\textbf{41.75}$^{+1.50}$\\ \bottomrule	
\end{tabular}
}
\label{table:ood}
\end{table*}

\begin{table}[t]
\setlength{\abovecaptionskip}{0cm}
\setlength{\belowcaptionskip}{0cm}
\centering
\caption{IID testing performance  (\%) comparison among different methods on MOSEI and MOSI datasets. \textit{Acc-2} and \textit{F1} are calculated as ``negative/non-negative''. We omitted the similar results of ``negative/positive'' to save space.
}
\setlength{\tabcolsep}{2mm}{
\resizebox{0.48\textwidth}{!}{
\begin{tabular}{c|cc|c|cc|c}
\toprule
\multicolumn{1}{c|}{\multirow{3}{*}{Model}} & \multicolumn{3}{c|}{MOSEI} & \multicolumn{3}{c}{MOSI}  \\
& \multicolumn{2}{c}{2-class} & \multicolumn{1}{c|}{7-class} & \multicolumn{2}{c}{2-class} & \multicolumn{1}{c}{7-class}  \\ \cline{2-7} 
& \textit{Acc-2} & \textit{F1} & \textit{Acc-7}& \textit{Acc-2} & \textit{F1} &  \textit{Acc-7}  \\ \hline
TFN	& 81.59 & 81.54   &52.11 &80.24 & 80.31 & 40.07  \\ 
LMF	&79.59 & 80.34  	&48.27 &79.85 & 79.95 &35.04  \\ 
MulT &81.05 &81.44 &	53.21	&79.61 &79.71 &35.19	\\ \hline
MAG-BERT	& 82.82 & 83.19 	&53.52	& 83.91 & 83.96	& 46.97 \\
$\quad$+CLUE (Ours) &\textbf{84.62} &\textbf{85.46}	&\textbf{53.68}	&\textbf{84.37} & \textbf{84.28} & \textbf{48.84} \\ \hline 
MISA & 82.17 & 82.61	&	\textbf{53.26}&	83.52& 83.58 &	{45.26}\\ 
$\quad$+CLUE (Ours)	&\textbf{84.51} & \textbf{85.28}&53.15	&\textbf{84.07} &\textbf{84.16} &\textbf{46.31}	 \\ \hline
Self-MM &83.71& 83.80	&53.31 &84.14&84.17	&\textbf{48.74}\\ 
$\quad$ +CLUE (Ours) &\textbf{84.52} &\textbf{84.46} 	&\textbf{53.42}		&\textbf{84.31} & \textbf{84.38} &	{48.04}\\ \bottomrule
\end{tabular}
}
}
\label{table:iid}
\end{table}

To split the datasets, we first divided each dataset (\ie MOSI or MOSEI) into two parts: the IID set, and the OOD set, where the former can be further split into three chunks (\ie IID training, IID validation, and IID testing) with random sampling, and the latter constitutes the OOD testing set. 
To achieve this, we adapted the simulated annealing algorithm, which is originally designed for solving the traveling salesman problem (TSP), for the dataset split. 
The adapted simulated annealing algorithm (as shown in Algorithm \ref{alg:ood}) will iterate until the overall distribution difference of all words over different sentiment categories between the IID set and OOD set gets sufficiently close to the pre-defined  distribution difference $\boldsymbol{\phi}_{\Delta}$. 
The number of iterations is set to $800$. The temperature parameter of this algorithm is set to $0.5$, and the temperature decay is set to $0.99$.  Ultimately, the OOD testing set derived from MOSI contains $12$ videos. To balance the size of OOD testing set and IID testing set, we randomly selected $12$ videos from the IID set obtained by the above algorithm as the IID testing set, while the rest is divided into two groups: $59$ ($85\%$) for IID training and  $10$ ($15\%$) for IID validation. Similarly, the IID training, IID validation, IID testing, OOD testing sets obtained from  MOSEI contain $1,830$, $324$, $330$, and $330$ videos, respectively.  
Towards intuitive understanding, we visualized the distributions of the frequent words in both the IID training set and the OOD testing set derived from MOSEI in Figure~\ref{fig:ood}. 
As can be seen, their distributions are significantly different to each other.

\subsubsection{\textbf{Baselines.}}
To evaluate the performance of CLUE, we adopted the following methods for comparison.

(1) \textbf{TFN}. The Tensor Fusion Network~\cite{tfn} learns both the intra-modality and inter-modality dynamics via Tensor Fusion and Modality Embedding Subnetworks, respectively. 

(2) \textbf{LMF}. Low-rank Multimodal Fusion~\cite{lmf} performs multimodal fusion using modality-specific low-rank tensors, which decreases the computational complexity and improves the model's efficiency.

(3) \textbf{MulT}. The core of Multimodal Transformer~\cite{mult} is the cross-modal attention mechanism, which directly attends to low-level features in other modalities. These interactions lead to better performance on unaligned multimodal datasets.

(4) \textbf{MAG-BERT}. The Multimodal Adaptation Gate~\cite{mag} regards the nonverbal behavior as a vector with a trajectory and magnitude, which is subsequently used to shift lexical representations within the pre-trained transformer model that demonstrates its advancement in many natural language tasks~\cite{v2p,bert}. 

(5) \textbf{MISA}. By projecting each modality of samples into two subspaces, this method learns both modality-invariant and -specific representations~\cite{DBLP:conf/mm/HazarikaZP20}, which then are fused for sentiment analysis. 

(6) \textbf{Self-MM}. This method introduces a unimodal label generation strategy based on the self-supervised method~\cite{selfmm}. Furthermore, a novel weight
self-adjusting strategy is introduced to balance different
task loss constraints.
It is worth mentioning that all the baselines treat the task of MSA as a regression task, which can obtain more supervisory information, while our CLUE treats MSA as a classification task.

\begin{table}[t]
\setlength{\abovecaptionskip}{0cm}
\setlength{\belowcaptionskip}{0cm}
\centering
\caption{Ablation study results (\%) for the binary classification (negative/non-negative) of our proposed CLUE on MOSEI. 
The best results are highlighted in boldface. 
}

\setlength{\tabcolsep}{5mm}{
\resizebox{0.48\textwidth}{!}{
\begin{tabular}{l|cc|cc}
\toprule 
\multicolumn{1}{c|}{\multirow{2}{*}{Model}}    
& \multicolumn{2}{c|}{IID testing} & \multicolumn{2}{c}{OOD testing}\\ 
& \textit{Acc-2} & \textit{F1} & \textit{Acc-2} & \textit{F1} \\ \hline

MAG-BERT+CLUE    & 84.62 & 85.46    &\textbf{78.34}    &\textbf{78.23}\\ 
$\quad$w/o-MSA model &   80.31  & 81.52 & 65.49  & 68.01\\ 
$\quad$w/o-text model       &\textbf{85.09}    &\textbf{85.60}    &74.37  &74.83\\ 
$\quad$w/o-KL loss   &{84.55}    &84.45    &78.28    &78.17\\ \hline

MISA+CLUE	  &84.51	  &85.28	  &  \textbf{77.17} & \textbf{77.08} \\ 
$\quad$w/o-MSA model       &80.31    &81.52   &65.49    &68.01\\ 
$\quad$w/o-text model      &84.31    &85.26    &74.53    &75.78\\
$\quad$w/o-KL loss &\textbf{84.69} &\textbf{85.44} & 76.75 &76.77\\ \hline

Self-MM+CLUE    &84.52    &84.46    &\textbf{77.76}    &\textbf{77.72}\\ 
$\quad$w/o-MSA model &80.31    &81.52  &65.49   &68.01\\ 
$\quad$w/o-text model       &84.52    &\textbf{85.41}    &73.51  &74.12\\ 
$\quad$w/o-KL loss   &\textbf{84.63}    &84.53    &77.41    &77.43\\

\bottomrule

\end{tabular}}}
\vspace{-0.3cm}
\label{ablation}
\end{table}

\begin{figure*}[t]
\setlength{\abovecaptionskip}{-0cm}
\setlength{\belowcaptionskip}{-0.30cm}
    \centering
    \includegraphics[scale=0.58]{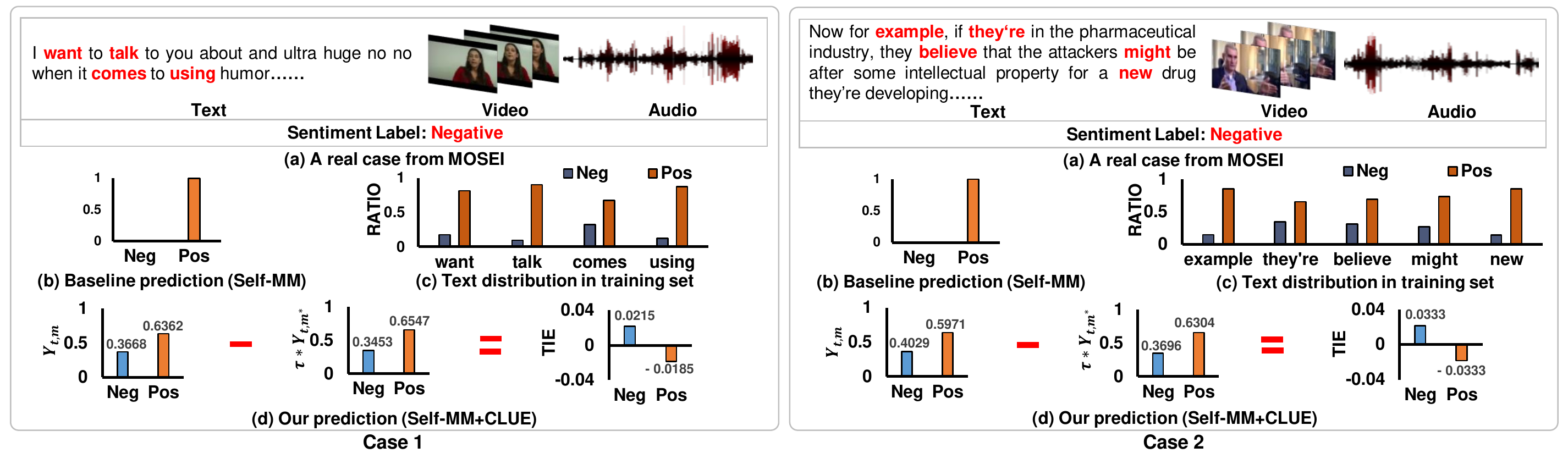}
    \caption{Cases of the binary sentiment classification (\ie positive (Pos) and negative (Neg) categories) by self-MM and CLUE. RATIO: the proportion of samples with the target words in the positive or negative categories of the training set.}
    \label{fig:case}
\end{figure*}

\vspace{-0.2cm}

\subsubsection{\textbf{Implementation Details.}}
In this work, we adopted three state-of-the-art (SOTA) MSA models (\ie MISA, Self-MM, and MAG-BERT) as our MSA backbone of our CLUE framework, respectively. 
We use audio and video features provided by the MOSI and MOSEI followed the three MSA models.
Regarding text modality, we also adopted the same setting with each of the three MSA models, where the pre-trained BERT is used and whether it should be fine-tuned depends on the original settings of these MSA models. 
Pertaining to the input of our Text Model branch, we simply adopted the 300D Glove~\cite{JeffreyPennington2014GloveGV} vectors as the word embeddings for our Text Model to better verify the effectiveness of our framework. Theoretically, the performance would be better if we use BERT embeddings.
Regarding optimization, we adopted the Adam as the optimizer with the learning rate of $3e-5$ for our Text Model and learning rates of $6e-5$, $2e-5$, and $1e-5$ for different MSA backbones (\ie MISA, Self-MM, and MAG-BERT). The dropout rates for these three MSA backbones are set to $0.1$, $0.3$, and $0.1$, respectively. The batchsize is set to $24$ for MOSEI dataset, and $32$ for MOSI dataset. 
We used the grid search strategy to find the optimal hyperparameters. Specifically, the hyperparameters $\alpha$, $\beta$, $\gamma$ are searched within $[0, 1]$ with the step size of $0.1$, while the hyperparameter $\tau$ is searched within $[0, 2.5]$ with the corresponding step size of $0.15$.

\subsection{On Model Comparison (RQ1 \& RQ2)}
We adopted the SOTA MAG-BERT, MISA and Self-MM as the MSA backbone of our method, respectively.
Table~\ref{table:ood} and Table~\ref{table:iid} show the performance comparison among different methods over OOD and IID testing settings, respectively. From these two tables, we have the following observations.
1) The models with our CLUE framework significantly outperform their original versions in the OOD testing setting. 
In particular, MAG-BERT+CLUE raises the \textit{Acc-2} of MAG-BERT from $76.41\%$ to $80.51\%$ for negative/positive on MOSEI. 
This validates the superior generalization ability of our CLUE framework over existing methods. 2) Besides, even in the IID testing, our MAG-BERT+CLUE also achieves the best performance in terms of most evaluation metrics on both datasets. And 3) for all the methods, their performance in the IID testing setting is better than that in the OOD testing setting. This confirms that 
the spurious correlations between textual words and sentiment labels do hurt the generalization ability of the model. However, we noticed that the performance decrease of our model is the least compared with that of the baseline methods. This reflects that our  CLUE is more robust for the OOD testing.





\subsection{On Ablation Study (RQ3)}
To verify the effect of the MSA model branch, the text model branch, as well as the KL loss for regularizing $Y_{t,m}$ and $Y_{t,m^*}$, we introduced three derivatives of CLUE: \textbf{w/o-MSA model},  \textbf{w/o-text model}, and  \textbf{w/o-KL loss}, by setting $\alpha=0$ in Eqn. ($\ref{eq:celoss}$), $\beta=0$ in Eqn. ($\ref{eq:celoss}$), and $\gamma = 0$ in Eqn. ($\ref{eqn:allloss}$), respectively.
Table~\ref{ablation} shows the ablation study results of our proposed CLUE on the MOSEI dataset with the task of binary classification. As can be seen, for IID testing, our MAG-BERT+CLUE, MISA+CLUE and Self-MM+CLUE achieve similar performance with their corresponding derivatives. This may be due to the fact that or because the bias is similar in training and testing set under IID setting. As for OOD testing, we have the following observations: 1) Our model largely  outperforms the derivative w/o-MSA model with different MSA backbones (\ie MAG-BERT, MISA and Self-MM). This implies the necessity of introducing the multi-modal content of videos towards sentiment analysis. 
2) For each backbone, our model exceeds the derivative w/o-text model across different evaluation metrics. This demonstrates the importance of mitigating the bad effect of textual modality with causal inference. 
And 3) both intact models surpass their w/o-KL loss derivatives. This shows that a proper $Y_{m^*}$  is crucial for optimal performance. 

\subsection{On Case Study}
To get an intuitive understanding of the generalization of our model, we illustrated  the binary (\textit{negative/positive}) classification results: Self-MM
and our CLUE, which adopts Self-MM as the MSA model backbone, 
on two testing samples from MOSEI dataset in Figure~\ref{fig:case}. To gain deeper insights, we provided the distribution of the keywords mentioned in the sample's text, highlighted in red, over the positive and negative categories in the training set, as well as the classification results of each branch of our model.
As can be seen, for both samples, our CLUE gives the correct sentiment, while Self-MM fails. Taking the Case 1 as an example, we can see that each branch of our model gives a higher probability to the positive category, so does the baseline Self-MM. However, combining the two branches with the causal inference, our model gives the negative prediction. This may be due to the fact that the keywords (\eg ``example'', ``they're'', and ``believe'') of the given testing sample in the training set are highly related to the positive sentiment. Thereby, the spurious correlation between words and sentiment labels mislead the two branches and the baseline Self-MM to predict the correlation rather than the causal relationship. 
This intuitively demonstrates the necessity of considering the spurious correlations 
 between the text and sentiment. 
Similar observation can be found in Case 2. 

%% file: 5_conclusion.tex
\vspace{-0.3cm}
\section{Conclusion and Future Work}
In this work, we first analyzed the spurious correlations between textual modality and the sentiment labels, and then defined the OOD MSA task to evaluate the generalization ability of MSA models. 
Besides, we presented a model-agnostic counterfactual multimodal sentiment analysis framework that can retain the good effect of textual modality and mitigate its bad effect. 
Extensive experiments on the two benchmarks demonstrate the superiority of the proposed framework on both IID and OOD testing. 

This work makes the initial step to consider OOD generalization in the MSA task, leaving many promising directions to future work. 
1) As an initial work on the OOD MSA, we focus on the most harmful spurious correlations in textual modality. It is meaningful to explore the bad correlations in other modalities. 
2) We adopt counterfactual reasoning to mitigate the direct effect of texts. To further improve the OOD generalization ability, another future direction is to discover more fine-grained causal relationships, \eg word-level correlation discovery. 
And 3) future work can study more strategies to implement the CLUE framework, such as superior fusion functions and text models.


%% file: main.bbl

\begin{thebibliography}{45}


\ifx \showCODEN    \undefined \def \showCODEN     #1{\unskip}     \fi
\ifx \showDOI      \undefined \def \showDOI       #1{#1}\fi
\ifx \showISBNx    \undefined \def \showISBNx     #1{\unskip}     \fi
\ifx \showISBNxiii \undefined \def \showISBNxiii  #1{\unskip}     \fi
\ifx \showISSN     \undefined \def \showISSN      #1{\unskip}     \fi
\ifx \showLCCN     \undefined \def \showLCCN      #1{\unskip}     \fi
\ifx \shownote     \undefined \def \shownote      #1{#1}          \fi
\ifx \showarticletitle \undefined \def \showarticletitle #1{#1}   \fi
\ifx \showURL      \undefined \def \showURL       {\relax}        \fi
\providecommand\bibfield[2]{#2}
\providecommand\bibinfo[2]{#2}
\providecommand\natexlab[1]{#1}
\providecommand\showeprint[2][]{arXiv:#2}

\bibitem[\protect\citeauthoryear{Andrew, Arora, Bilmes, and Livescu}{Andrew
  et~al\mbox{.}}{2013}]%
        {correlation}
\bibfield{author}{\bibinfo{person}{Galen Andrew}, \bibinfo{person}{Raman
  Arora}, \bibinfo{person}{Jeff~A. Bilmes}, {and} \bibinfo{person}{Karen
  Livescu}.} \bibinfo{year}{2013}\natexlab{}.
\newblock \showarticletitle{Deep Canonical Correlation Analysis}. In
  \bibinfo{booktitle}{\emph{Proceedings of the International Conference on
  Machine Learning}}. \bibinfo{publisher}{JMLR.org},
  \bibinfo{pages}{1247--1255}.
\newblock


\bibitem[\protect\citeauthoryear{Devlin, Chang, Lee, and Toutanova}{Devlin
  et~al\mbox{.}}{2019}]%
        {bert}
\bibfield{author}{\bibinfo{person}{Jacob Devlin}, \bibinfo{person}{Ming{-}Wei
  Chang}, \bibinfo{person}{Kenton Lee}, {and} \bibinfo{person}{Kristina
  Toutanova}.} \bibinfo{year}{2019}\natexlab{}.
\newblock \showarticletitle{{BERT:} Pre-training of Deep Bidirectional
  Transformers for Language Understanding}. In
  \bibinfo{booktitle}{\emph{Proceedings of the Annual Meeting of the
  Association for Computational Linguistics}}. \bibinfo{publisher}{{ACL}},
  \bibinfo{pages}{4171--4186}.
\newblock


\bibitem[\protect\citeauthoryear{Dobri{\v{s}}ek, Gaj{\v{s}}ek, Miheli{\v{c}},
  Pave{\v{s}}i{\'c}, and {\v{S}}truc}{Dobri{\v{s}}ek et~al\mbox{.}}{2013}]%
        {latefusion}
\bibfield{author}{\bibinfo{person}{Simon Dobri{\v{s}}ek}, \bibinfo{person}{Rok
  Gaj{\v{s}}ek}, \bibinfo{person}{France Miheli{\v{c}}},
  \bibinfo{person}{Nikola Pave{\v{s}}i{\'c}}, {and} \bibinfo{person}{Vitomir
  {\v{S}}truc}.} \bibinfo{year}{2013}\natexlab{}.
\newblock \showarticletitle{Towards efficient multi-modal emotion recognition}.
\newblock \bibinfo{journal}{\emph{International Journal of Advanced Robotic
  Systems}} \bibinfo{volume}{10}, \bibinfo{number}{1} (\bibinfo{year}{2013}),
  \bibinfo{pages}{53}.
\newblock


\bibitem[\protect\citeauthoryear{Feng, Zhang, He, Zhang, and Chua}{Feng
  et~al\mbox{.}}{2021}]%
        {lanunder}
\bibfield{author}{\bibinfo{person}{Fuli Feng}, \bibinfo{person}{Jizhi Zhang},
  \bibinfo{person}{Xiangnan He}, \bibinfo{person}{Hanwang Zhang}, {and}
  \bibinfo{person}{Tat{-}Seng Chua}.} \bibinfo{year}{2021}\natexlab{}.
\newblock \showarticletitle{Empowering Language Understanding with
  Counterfactual Reasoning}. In \bibinfo{booktitle}{\emph{Proceedings of the
  Annual Meeting of the Association for Computational Linguistics}}.
  \bibinfo{publisher}{{ACL}}, \bibinfo{pages}{2226--2236}.
\newblock


\bibitem[\protect\citeauthoryear{Geirhos, Jacobsen, Michaelis, Zemel, Brendel,
  Bethge, and Wichmann}{Geirhos et~al\mbox{.}}{2020}]%
        {DBLP:journals/natmi/GeirhosJMZBBW20}
\bibfield{author}{\bibinfo{person}{Robert Geirhos},
  \bibinfo{person}{J{\"{o}}rn{-}Henrik Jacobsen}, \bibinfo{person}{Claudio
  Michaelis}, \bibinfo{person}{Richard~S. Zemel}, \bibinfo{person}{Wieland
  Brendel}, \bibinfo{person}{Matthias Bethge}, {and} \bibinfo{person}{Felix~A.
  Wichmann}.} \bibinfo{year}{2020}\natexlab{}.
\newblock \showarticletitle{Shortcut learning in deep neural networks}.
\newblock \bibinfo{journal}{\emph{Nature Machine Intelligence}}
  \bibinfo{volume}{2}, \bibinfo{number}{11} (\bibinfo{year}{2020}),
  \bibinfo{pages}{665--673}.
\newblock


\bibitem[\protect\citeauthoryear{Glymour, Pearl, and Jewell}{Glymour
  et~al\mbox{.}}{2016}]%
        {glymour2016causal}
\bibfield{author}{\bibinfo{person}{Madelyn Glymour}, \bibinfo{person}{Judea
  Pearl}, {and} \bibinfo{person}{Nicholas~P Jewell}.}
  \bibinfo{year}{2016}\natexlab{}.
\newblock \bibinfo{booktitle}{\emph{Causal inference in statistics: A primer}}.
\newblock \bibinfo{publisher}{John Wiley \& Sons}.
\newblock


\bibitem[\protect\citeauthoryear{Gupta, Mehrotra, Bansal, and Kumari}{Gupta
  et~al\mbox{.}}{2017}]%
        {bayes}
\bibfield{author}{\bibinfo{person}{Paridhi Gupta}, \bibinfo{person}{Tanu
  Mehrotra}, \bibinfo{person}{Ashita Bansal}, {and} \bibinfo{person}{Bhawna
  Kumari}.} \bibinfo{year}{2017}\natexlab{}.
\newblock \showarticletitle{Multimodal sentiment analysis and context
  determination: Using perplexed Bayes classification}. In
  \bibinfo{booktitle}{\emph{International Conference on Automation and
  Computing}}. \bibinfo{publisher}{{IEEE}}, \bibinfo{pages}{1--6}.
\newblock


\bibitem[\protect\citeauthoryear{Hazarika, Zimmermann, and Poria}{Hazarika
  et~al\mbox{.}}{2020}]%
        {DBLP:conf/mm/HazarikaZP20}
\bibfield{author}{\bibinfo{person}{Devamanyu Hazarika}, \bibinfo{person}{Roger
  Zimmermann}, {and} \bibinfo{person}{Soujanya Poria}.}
  \bibinfo{year}{2020}\natexlab{}.
\newblock \showarticletitle{{MISA:} Modality-Invariant and -Specific
  Representations for Multimodal Sentiment Analysis}. In
  \bibinfo{booktitle}{\emph{Proceedings of the {ACM} International Conference
  on Multimedia}}. \bibinfo{publisher}{{ACM}}, \bibinfo{pages}{1122--1131}.
\newblock


\bibitem[\protect\citeauthoryear{Hu, Qian, Fang, Wang, Zhao, Zhang, and Xu}{Hu
  et~al\mbox{.}}{2021}]%
        {hu2021efficient}
\bibfield{author}{\bibinfo{person}{Jun Hu}, \bibinfo{person}{Shengsheng Qian},
  \bibinfo{person}{Quan Fang}, \bibinfo{person}{Youze Wang},
  \bibinfo{person}{Quan Zhao}, \bibinfo{person}{Huaiwen Zhang}, {and}
  \bibinfo{person}{Changsheng Xu}.} \bibinfo{year}{2021}\natexlab{}.
\newblock \showarticletitle{Efficient graph deep learning in tensorflow with
  tf\_geometric}. In \bibinfo{booktitle}{\emph{Proceedings of the ACM
  International Conference on Multimedia}}. \bibinfo{publisher}{ACM},
  \bibinfo{pages}{3775--3778}.
\newblock


\bibitem[\protect\citeauthoryear{Hu, Qian, Fang, and Xu}{Hu
  et~al\mbox{.}}{2019}]%
        {hu2019hierarchical}
\bibfield{author}{\bibinfo{person}{Jun Hu}, \bibinfo{person}{Shengsheng Qian},
  \bibinfo{person}{Quan Fang}, {and} \bibinfo{person}{Changsheng Xu}.}
  \bibinfo{year}{2019}\natexlab{}.
\newblock \showarticletitle{Hierarchical graph semantic pooling network for
  multi-modal community question answer matching}. In
  \bibinfo{booktitle}{\emph{Proceedings of the ACM International Conference on
  Multimedia}}. \bibinfo{publisher}{ACM}, \bibinfo{pages}{1157--1165}.
\newblock


\bibitem[\protect\citeauthoryear{Huddar, Sannakki, and Rajpurohit}{Huddar
  et~al\mbox{.}}{2018}]%
        {RF}
\bibfield{author}{\bibinfo{person}{Mahesh~G Huddar}, \bibinfo{person}{Sanjeev~S
  Sannakki}, {and} \bibinfo{person}{Vijay~S Rajpurohit}.}
  \bibinfo{year}{2018}\natexlab{}.
\newblock \showarticletitle{An ensemble approach to utterance level multimodal
  sentiment analysis}. In \bibinfo{booktitle}{\emph{International Conference on
  Computational Techniques, Electronics and Mechanical Systems}}.
  \bibinfo{publisher}{{IEEE}}, \bibinfo{pages}{145--150}.
\newblock


\bibitem[\protect\citeauthoryear{Khan and Fu}{Khan and Fu}{2021}]%
        {DBLP:conf/mm/0001F21}
\bibfield{author}{\bibinfo{person}{Zaid Khan} {and} \bibinfo{person}{Yun Fu}.}
  \bibinfo{year}{2021}\natexlab{}.
\newblock \showarticletitle{Exploiting {BERT} for Multimodal Target Sentiment
  Classification through Input Space Translation}. In
  \bibinfo{booktitle}{\emph{Proceedings of the {ACM} International Conference
  on Multimedia}}. \bibinfo{publisher}{{ACM}}, \bibinfo{pages}{3034--3042}.
\newblock


\bibitem[\protect\citeauthoryear{Li, Wang, Xiao, Ji, and Chua}{Li
  et~al\mbox{.}}{2022}]%
        {li2022invariant}
\bibfield{author}{\bibinfo{person}{Yicong Li}, \bibinfo{person}{Xiang Wang},
  \bibinfo{person}{Junbin Xiao}, \bibinfo{person}{Wei Ji}, {and}
  \bibinfo{person}{Tat-Seng Chua}.} \bibinfo{year}{2022}\natexlab{}.
\newblock \showarticletitle{Invariant grounding for video question answering}.
  In \bibinfo{booktitle}{\emph{Proceedings of the IEEE/CVF Conference on
  Computer Vision and Pattern Recognition}}. \bibinfo{publisher}{IEEE},
  \bibinfo{pages}{2928--2937}.
\newblock


\bibitem[\protect\citeauthoryear{Li, Yang, Shang, and Chua}{Li
  et~al\mbox{.}}{2021}]%
        {li2021interventional}
\bibfield{author}{\bibinfo{person}{Yicong Li}, \bibinfo{person}{Xun Yang},
  \bibinfo{person}{Xindi Shang}, {and} \bibinfo{person}{Tat-Seng Chua}.}
  \bibinfo{year}{2021}\natexlab{}.
\newblock \showarticletitle{Interventional video relation detection}. In
  \bibinfo{booktitle}{\emph{Proceedings of the ACM International Conference on
  Multimedia}}. \bibinfo{publisher}{ACM}, \bibinfo{pages}{4091--4099}.
\newblock


\bibitem[\protect\citeauthoryear{Liu, Cheng, Zhu, Gao, and Nie}{Liu
  et~al\mbox{.}}{2021}]%
        {Liu2021IMP_GCN}
\bibfield{author}{\bibinfo{person}{Fan Liu}, \bibinfo{person}{Zhiyong Cheng},
  \bibinfo{person}{Lei Zhu}, \bibinfo{person}{Zan Gao}, {and}
  \bibinfo{person}{Liqiang Nie}.} \bibinfo{year}{2021}\natexlab{}.
\newblock \showarticletitle{Interest-Aware Message-Passing GCN for
  Recommendation}. In \bibinfo{booktitle}{\emph{Proceedings of the Web
  Conference}}. \bibinfo{publisher}{ACM}, \bibinfo{pages}{1296–1305}.
\newblock


\bibitem[\protect\citeauthoryear{Liu, Shen, Lakshminarasimhan, Liang, Zadeh,
  and Morency}{Liu et~al\mbox{.}}{2018}]%
        {lmf}
\bibfield{author}{\bibinfo{person}{Zhun Liu}, \bibinfo{person}{Ying Shen},
  \bibinfo{person}{Varun~Bharadhwaj Lakshminarasimhan},
  \bibinfo{person}{Paul~Pu Liang}, \bibinfo{person}{Amir Zadeh}, {and}
  \bibinfo{person}{Louis{-}Philippe Morency}.} \bibinfo{year}{2018}\natexlab{}.
\newblock \showarticletitle{Efficient Low-rank Multimodal Fusion With
  Modality-Specific Factors}. In \bibinfo{booktitle}{\emph{Proceedings of the
  Annual Meeting of the Association for Computational Linguistics}}.
  \bibinfo{publisher}{{ACL}}, \bibinfo{pages}{2247--2256}.
\newblock


\bibitem[\protect\citeauthoryear{Mai, Hu, and Xing}{Mai et~al\mbox{.}}{2020}]%
        {autoencoder}
\bibfield{author}{\bibinfo{person}{Sijie Mai}, \bibinfo{person}{Haifeng Hu},
  {and} \bibinfo{person}{Songlong Xing}.} \bibinfo{year}{2020}\natexlab{}.
\newblock \showarticletitle{Modality to Modality Translation: An Adversarial
  Representation Learning and Graph Fusion Network for Multimodal Fusion}. In
  \bibinfo{booktitle}{\emph{Proceedings of the Conference on Artificial
  Intelligence}}. \bibinfo{publisher}{{AAAI}}, \bibinfo{pages}{164--172}.
\newblock


\bibitem[\protect\citeauthoryear{Ni, Song, Zhu, Zheng, and Gao}{Ni
  et~al\mbox{.}}{2021}]%
        {DBLP:conf/mm/NiSZZG21}
\bibfield{author}{\bibinfo{person}{Hao Ni}, \bibinfo{person}{Jingkuan Song},
  \bibinfo{person}{Xiaosu Zhu}, \bibinfo{person}{Feng Zheng}, {and}
  \bibinfo{person}{Lianli Gao}.} \bibinfo{year}{2021}\natexlab{}.
\newblock \showarticletitle{Camera-Agnostic Person Re-Identification via
  Adversarial Disentangling Learning}. In \bibinfo{booktitle}{\emph{The {ACM}
  International Conference on Multimedia}}. \bibinfo{publisher}{{ACM}},
  \bibinfo{pages}{2002--2010}.
\newblock


\bibitem[\protect\citeauthoryear{Niu, Tang, Zhang, Lu, Hua, and Wen}{Niu
  et~al\mbox{.}}{2021}]%
        {niu2021counterfactual}
\bibfield{author}{\bibinfo{person}{Yulei Niu}, \bibinfo{person}{Kaihua Tang},
  \bibinfo{person}{Hanwang Zhang}, \bibinfo{person}{Zhiwu Lu},
  \bibinfo{person}{Xiansheng Hua}, {and} \bibinfo{person}{Jirong Wen}.}
  \bibinfo{year}{2021}\natexlab{}.
\newblock \showarticletitle{Counterfactual {VQA:} {A} Cause-Effect Look at
  Language Bias}. In \bibinfo{booktitle}{\emph{{IEEE} Conference on Computer
  Vision and Pattern Recognition}}. \bibinfo{publisher}{{IEEE}},
  \bibinfo{pages}{12700--12710}.
\newblock


\bibitem[\protect\citeauthoryear{Pearl}{Pearl}{2009}]%
        {causality}
\bibfield{author}{\bibinfo{person}{Judea Pearl}.}
  \bibinfo{year}{2009}\natexlab{}.
\newblock \bibinfo{booktitle}{\emph{Causality}}.
\newblock \bibinfo{publisher}{Cambridge university press}.
\newblock


\bibitem[\protect\citeauthoryear{Pennington, Socher, and Manning}{Pennington
  et~al\mbox{.}}{2014}]%
        {JeffreyPennington2014GloveGV}
\bibfield{author}{\bibinfo{person}{Jeffrey Pennington},
  \bibinfo{person}{Richard Socher}, {and} \bibinfo{person}{Christopher~D.
  Manning}.} \bibinfo{year}{2014}\natexlab{}.
\newblock \showarticletitle{Glove: Global Vectors for Word Representation}. In
  \bibinfo{booktitle}{\emph{Proceedings of the Conference on Empirical Methods
  in Natural Language Processing}}. \bibinfo{publisher}{{ACL}},
  \bibinfo{pages}{1532--1543}.
\newblock


\bibitem[\protect\citeauthoryear{Peris and Casacuberta}{Peris and
  Casacuberta}{2019}]%
        {seq2seq}
\bibfield{author}{\bibinfo{person}{{\'{A}}lvaro Peris} {and}
  \bibinfo{person}{Francisco Casacuberta}.} \bibinfo{year}{2019}\natexlab{}.
\newblock \showarticletitle{A Neural, Interactive-predictive System for
  Multimodal Sequence to Sequence Tasks}. In
  \bibinfo{booktitle}{\emph{Proceedings of the Annual Meeting of the
  Association for Computational Linguistics}}. \bibinfo{publisher}{{ACL}},
  \bibinfo{pages}{81--86}.
\newblock


\bibitem[\protect\citeauthoryear{Pham, Liang, Manzini, Morency, and
  P{\'{o}}czos}{Pham et~al\mbox{.}}{2019}]%
        {cyclic}
\bibfield{author}{\bibinfo{person}{Hai Pham}, \bibinfo{person}{Paul~Pu Liang},
  \bibinfo{person}{Thomas Manzini}, \bibinfo{person}{Louis{-}Philippe Morency},
  {and} \bibinfo{person}{Barnab{\'{a}}s P{\'{o}}czos}.}
  \bibinfo{year}{2019}\natexlab{}.
\newblock \showarticletitle{Found in Translation: Learning Robust Joint
  Representations by Cyclic Translations between Modalities}. In
  \bibinfo{booktitle}{\emph{Proceedings of the Conference on Artificial
  Intelligence}}. \bibinfo{publisher}{{AAAI}}, \bibinfo{pages}{6892--6899}.
\newblock


\bibitem[\protect\citeauthoryear{Qian, Feng, Wen, Ma, and Xie}{Qian
  et~al\mbox{.}}{2021}]%
        {qian2021counterfactual}
\bibfield{author}{\bibinfo{person}{Chen Qian}, \bibinfo{person}{Fuli Feng},
  \bibinfo{person}{Lijie Wen}, \bibinfo{person}{Chunping Ma}, {and}
  \bibinfo{person}{Pengjun Xie}.} \bibinfo{year}{2021}\natexlab{}.
\newblock \showarticletitle{Counterfactual Inference for Text Classification
  Debiasing}. In \bibinfo{booktitle}{\emph{Proceedings of the Annual Meeting of
  the Association for Computational Linguistics}}. \bibinfo{publisher}{{ACL}},
  \bibinfo{pages}{5434--5445}.
\newblock


\bibitem[\protect\citeauthoryear{Rahman, Hasan, Lee, Zadeh, Mao, Morency, and
  Hoque}{Rahman et~al\mbox{.}}{2020}]%
        {mag}
\bibfield{author}{\bibinfo{person}{Wasifur Rahman}, \bibinfo{person}{Md.~Kamrul
  Hasan}, \bibinfo{person}{Sangwu Lee}, \bibinfo{person}{AmirAli~Bagher Zadeh},
  \bibinfo{person}{Chengfeng Mao}, \bibinfo{person}{Louis{-}Philippe Morency},
  {and} \bibinfo{person}{Mohammed~E. Hoque}.} \bibinfo{year}{2020}\natexlab{}.
\newblock \showarticletitle{Integrating Multimodal Information in Large
  Pretrained Transformers}. In \bibinfo{booktitle}{\emph{Proceedings of the
  Annual Meeting of the Association for Computational Linguistics}}.
  \bibinfo{publisher}{{ACL}}, \bibinfo{pages}{2359--2369}.
\newblock


\bibitem[\protect\citeauthoryear{Robins}{Robins}{1986}]%
        {1986A}
\bibfield{author}{\bibinfo{person}{J. Robins}.}
  \bibinfo{year}{1986}\natexlab{}.
\newblock \showarticletitle{A New Approach to Causal Inference in Mortality
  Studies with Sustained Exposure Period}.
\newblock \bibinfo{journal}{\emph{Mathematical Modelling}} \bibinfo{volume}{7},
  \bibinfo{number}{9–12} (\bibinfo{year}{1986}), \bibinfo{pages}{1393--1512}.
\newblock


\bibitem[\protect\citeauthoryear{Song, Jing, Lin, Zhao, Chen, and Nie}{Song
  et~al\mbox{.}}{2022}]%
        {v2p}
\bibfield{author}{\bibinfo{person}{Xuemeng Song}, \bibinfo{person}{Liqiang
  Jing}, \bibinfo{person}{Dengtian Lin}, \bibinfo{person}{Zhongzhou Zhao},
  \bibinfo{person}{Haiqing Chen}, {and} \bibinfo{person}{Liqiang Nie}.}
  \bibinfo{year}{2022}\natexlab{}.
\newblock \showarticletitle{{V2P:} Vision-to-Prompt based Multi-Modal Product
  Summary Generation}. In \bibinfo{booktitle}{\emph{The International
  Conference on Research and Development in Information Retrieval}}.
  \bibinfo{publisher}{{ACM}}, \bibinfo{pages}{992--1001}.
\newblock


\bibitem[\protect\citeauthoryear{Sun, Wang, Song, Feng, and Nie}{Sun
  et~al\mbox{.}}{2022}]%
        {DBLP:journals/tomccap/SunWSFN22}
\bibfield{author}{\bibinfo{person}{Teng Sun}, \bibinfo{person}{Chun Wang},
  \bibinfo{person}{Xuemeng Song}, \bibinfo{person}{Fuli Feng}, {and}
  \bibinfo{person}{Liqiang Nie}.} \bibinfo{year}{2022}\natexlab{}.
\newblock \showarticletitle{Response Generation by Jointly Modeling
  Personalized Linguistic Styles and Emotions}.
\newblock \bibinfo{journal}{\emph{{ACM} Trans. Multim. Comput. Commun. Appl.}}
  \bibinfo{volume}{18}, \bibinfo{number}{2} (\bibinfo{year}{2022}),
  \bibinfo{pages}{52:1--52:20}.
\newblock


\bibitem[\protect\citeauthoryear{Tsai, Bai, Liang, Kolter, Morency, and
  Salakhutdinov}{Tsai et~al\mbox{.}}{2019}]%
        {mult}
\bibfield{author}{\bibinfo{person}{Yao{-}Hung~Hubert Tsai},
  \bibinfo{person}{Shaojie Bai}, \bibinfo{person}{Paul~Pu Liang},
  \bibinfo{person}{J.~Zico Kolter}, \bibinfo{person}{Louis{-}Philippe Morency},
  {and} \bibinfo{person}{Ruslan Salakhutdinov}.}
  \bibinfo{year}{2019}\natexlab{}.
\newblock \showarticletitle{Multimodal Transformer for Unaligned Multimodal
  Language Sequences}. In \bibinfo{booktitle}{\emph{Proceedings of the Annual
  Meeting of the Association for Computational Linguistics}}.
  \bibinfo{publisher}{{ACL}}, \bibinfo{pages}{6558--6569}.
\newblock


\bibitem[\protect\citeauthoryear{van Laarhoven and Aarts}{van Laarhoven and
  Aarts}{1987}]%
        {simulated}
\bibfield{author}{\bibinfo{person}{Peter J.~M. van Laarhoven} {and}
  \bibinfo{person}{Emile H.~L. Aarts}.} \bibinfo{year}{1987}\natexlab{}.
\newblock \bibinfo{booktitle}{\emph{Simulated Annealing: Theory and
  Applications}}. \bibinfo{series}{Mathematics and Its Applications},
  Vol.~\bibinfo{volume}{37}.
\newblock \bibinfo{publisher}{Springer}. 7--15 pages.
\newblock


\bibitem[\protect\citeauthoryear{Wang, Feng, He, Zhang, and Chua}{Wang
  et~al\mbox{.}}{2021}]%
        {DBLP:conf/sigir/WangF0ZC21}
\bibfield{author}{\bibinfo{person}{Wenjie Wang}, \bibinfo{person}{Fuli Feng},
  \bibinfo{person}{Xiangnan He}, \bibinfo{person}{Hanwang Zhang}, {and}
  \bibinfo{person}{Tat{-}Seng Chua}.} \bibinfo{year}{2021}\natexlab{}.
\newblock \showarticletitle{Clicks can be Cheating: Counterfactual
  Recommendation for Mitigating Clickbait Issue}. In
  \bibinfo{booktitle}{\emph{The International Conference on Research and
  Development in Information Retrieval}}. \bibinfo{publisher}{{ACM}},
  \bibinfo{pages}{1288--1297}.
\newblock


\bibitem[\protect\citeauthoryear{Wang, Lin, Feng, He, Lin, and Chua}{Wang
  et~al\mbox{.}}{2022}]%
        {wang2022causal}
\bibfield{author}{\bibinfo{person}{Wenjie Wang}, \bibinfo{person}{Xinyu Lin},
  \bibinfo{person}{Fuli Feng}, \bibinfo{person}{Xiangnan He},
  \bibinfo{person}{Min Lin}, {and} \bibinfo{person}{Tat-Seng Chua}.}
  \bibinfo{year}{2022}\natexlab{}.
\newblock \showarticletitle{Causal Representation Learning for
  Out-of-Distribution Recommendation}. In \bibinfo{booktitle}{\emph{Proceedings
  of the ACM Web Conference}}. \bibinfo{publisher}{ACM},
  \bibinfo{pages}{3562--3571}.
\newblock


\bibitem[\protect\citeauthoryear{Wang, Wan, and Wan}{Wang
  et~al\mbox{.}}{2020}]%
        {trans}
\bibfield{author}{\bibinfo{person}{Zilong Wang}, \bibinfo{person}{Zhaohong
  Wan}, {and} \bibinfo{person}{Xiaojun Wan}.} \bibinfo{year}{2020}\natexlab{}.
\newblock \showarticletitle{TransModality: An End2End Fusion Method with
  Transformer for Multimodal Sentiment Analysis}. In
  \bibinfo{booktitle}{\emph{The Web Conference}}. \bibinfo{publisher}{{ACM}},
  \bibinfo{pages}{2514--2520}.
\newblock


\bibitem[\protect\citeauthoryear{Wei, Wang, Li, Nie, Li, Li, and Chua}{Wei
  et~al\mbox{.}}{2021}]%
        {wei2021contrastive}
\bibfield{author}{\bibinfo{person}{Yinwei Wei}, \bibinfo{person}{Xiang Wang},
  \bibinfo{person}{Qi Li}, \bibinfo{person}{Liqiang Nie}, \bibinfo{person}{Yan
  Li}, \bibinfo{person}{Xuanping Li}, {and} \bibinfo{person}{Tat-Seng Chua}.}
  \bibinfo{year}{2021}\natexlab{}.
\newblock \showarticletitle{Contrastive learning for cold-start
  recommendation}. In \bibinfo{booktitle}{\emph{Proceedings of the ACM
  International Conference on Multimedia}}. \bibinfo{publisher}{ACM},
  \bibinfo{pages}{5382--5390}.
\newblock


\bibitem[\protect\citeauthoryear{Wei, Wang, Nie, He, Hong, and Chua}{Wei
  et~al\mbox{.}}{2019}]%
        {wei2019mmgcn}
\bibfield{author}{\bibinfo{person}{Yinwei Wei}, \bibinfo{person}{Xiang Wang},
  \bibinfo{person}{Liqiang Nie}, \bibinfo{person}{Xiangnan He},
  \bibinfo{person}{Richang Hong}, {and} \bibinfo{person}{Tat-Seng Chua}.}
  \bibinfo{year}{2019}\natexlab{}.
\newblock \showarticletitle{MMGCN: Multi-modal graph convolution network for
  personalized recommendation of micro-video}. In
  \bibinfo{booktitle}{\emph{Proceedings of the ACM International Conference on
  Multimedia}}. \bibinfo{publisher}{ACM}, \bibinfo{pages}{1437--1445}.
\newblock


\bibitem[\protect\citeauthoryear{Yan, Zhao, and Lu}{Yan et~al\mbox{.}}{2021}]%
        {GAN}
\bibfield{author}{\bibinfo{person}{Xu Yan}, \bibinfo{person}{Li{-}Ming Zhao},
  {and} \bibinfo{person}{Bao{-}Liang Lu}.} \bibinfo{year}{2021}\natexlab{}.
\newblock \showarticletitle{Simplifying Multimodal Emotion Recognition with
  Single Eye Movement Modality}. In \bibinfo{booktitle}{\emph{Proceedings of
  the {ACM} International Conference on Multimedia}}.
  \bibinfo{publisher}{{ACM}}, \bibinfo{pages}{1057--1063}.
\newblock


\bibitem[\protect\citeauthoryear{Yu, Xu, Yuan, and Wu}{Yu
  et~al\mbox{.}}{2021}]%
        {selfmm}
\bibfield{author}{\bibinfo{person}{Wenmeng Yu}, \bibinfo{person}{Hua Xu},
  \bibinfo{person}{Ziqi Yuan}, {and} \bibinfo{person}{Jiele Wu}.}
  \bibinfo{year}{2021}\natexlab{}.
\newblock \showarticletitle{Learning Modality-Specific Representations with
  Self-Supervised Multi-Task Learning for Multimodal Sentiment Analysis}. In
  \bibinfo{booktitle}{\emph{Proceedings of the Conference on Artificial
  Intelligence}}. \bibinfo{publisher}{{AAAI}}, \bibinfo{pages}{10790--10797}.
\newblock


\bibitem[\protect\citeauthoryear{Yuan, Li, Xu, and Yu}{Yuan
  et~al\mbox{.}}{2021}]%
        {DBLP:conf/mm/YuanLXY21}
\bibfield{author}{\bibinfo{person}{Ziqi Yuan}, \bibinfo{person}{Wei Li},
  \bibinfo{person}{Hua Xu}, {and} \bibinfo{person}{Wenmeng Yu}.}
  \bibinfo{year}{2021}\natexlab{}.
\newblock \showarticletitle{Transformer-based Feature Reconstruction Network
  for Robust Multimodal Sentiment Analysis}. In \bibinfo{booktitle}{\emph{The
  {ACM} International Conference on Multimedia}}. \bibinfo{publisher}{{ACM}},
  \bibinfo{pages}{4400--4407}.
\newblock


\bibitem[\protect\citeauthoryear{Zadeh, Chen, Poria, Cambria, and
  Morency}{Zadeh et~al\mbox{.}}{2017}]%
        {tfn}
\bibfield{author}{\bibinfo{person}{Amir Zadeh}, \bibinfo{person}{Minghai Chen},
  \bibinfo{person}{Soujanya Poria}, \bibinfo{person}{Erik Cambria}, {and}
  \bibinfo{person}{Louis{-}Philippe Morency}.} \bibinfo{year}{2017}\natexlab{}.
\newblock \showarticletitle{Tensor Fusion Network for Multimodal Sentiment
  Analysis}. In \bibinfo{booktitle}{\emph{Proceedings of the Conference on
  Empirical Methods in Natural Language Processing}}. \bibinfo{publisher}{ACL},
  \bibinfo{pages}{1103--1114}.
\newblock


\bibitem[\protect\citeauthoryear{Zadeh, Liang, Poria, Cambria, and
  Morency}{Zadeh et~al\mbox{.}}{2018a}]%
        {DBLP:conf/acl/MorencyCPLZ18}
\bibfield{author}{\bibinfo{person}{Amir Zadeh}, \bibinfo{person}{Paul~Pu
  Liang}, \bibinfo{person}{Soujanya Poria}, \bibinfo{person}{Erik Cambria},
  {and} \bibinfo{person}{Louis{-}Philippe Morency}.}
  \bibinfo{year}{2018}\natexlab{a}.
\newblock \showarticletitle{Multimodal Language Analysis in the Wild:
  {CMU-MOSEI} Dataset and Interpretable Dynamic Fusion Graph}. In
  \bibinfo{booktitle}{\emph{Proceedings of the Annual Meeting of the
  Association for Computational Linguistics}}. \bibinfo{publisher}{{ACL}},
  \bibinfo{pages}{2236--2246}.
\newblock


\bibitem[\protect\citeauthoryear{Zadeh, Liang, Poria, Vij, Cambria, and
  Morency}{Zadeh et~al\mbox{.}}{2018b}]%
        {DBLP:conf/aaai/ZadehLPVCM18}
\bibfield{author}{\bibinfo{person}{Amir Zadeh}, \bibinfo{person}{Paul~Pu
  Liang}, \bibinfo{person}{Soujanya Poria}, \bibinfo{person}{Prateek Vij},
  \bibinfo{person}{Erik Cambria}, {and} \bibinfo{person}{Louis{-}Philippe
  Morency}.} \bibinfo{year}{2018}\natexlab{b}.
\newblock \showarticletitle{Multi-attention Recurrent Network for Human
  Communication Comprehension}. In \bibinfo{booktitle}{\emph{Proceedings of the
  Conference on Artificial Intelligence}}. \bibinfo{publisher}{{AAAI}},
  \bibinfo{pages}{5642--5649}.
\newblock


\bibitem[\protect\citeauthoryear{Zadeh, Zellers, Pincus, and Morency}{Zadeh
  et~al\mbox{.}}{2016}]%
        {DBLP:journals/corr/ZadehZPM16}
\bibfield{author}{\bibinfo{person}{Amir Zadeh}, \bibinfo{person}{Rowan
  Zellers}, \bibinfo{person}{Eli Pincus}, {and}
  \bibinfo{person}{Louis{-}Philippe Morency}.} \bibinfo{year}{2016}\natexlab{}.
\newblock \showarticletitle{{MOSI:} Multimodal Corpus of Sentiment Intensity
  and Subjectivity Analysis in Online Opinion Videos}.
\newblock \bibinfo{journal}{\emph{IEEE Intelligent Systems}}
  \bibinfo{volume}{36}, \bibinfo{number}{6} (\bibinfo{year}{2016}),
  \bibinfo{pages}{82--88}.
\newblock


\bibitem[\protect\citeauthoryear{Zhang, Li, Zhu, and Zhou}{Zhang
  et~al\mbox{.}}{2019}]%
        {DBLP:conf/mm/ZhangLZZ19}
\bibfield{author}{\bibinfo{person}{Dong Zhang}, \bibinfo{person}{Shoushan Li},
  \bibinfo{person}{Qiaoming Zhu}, {and} \bibinfo{person}{Guodong Zhou}.}
  \bibinfo{year}{2019}\natexlab{}.
\newblock \showarticletitle{Effective Sentiment-relevant Word Selection for
  Multi-modal Sentiment Analysis in Spoken Language}. In
  \bibinfo{booktitle}{\emph{Proceedings of the {ACM} International Conference
  on Multimedia}}. \bibinfo{publisher}{{ACM}}, \bibinfo{pages}{148--156}.
\newblock


\bibitem[\protect\citeauthoryear{Zhang, Zhang, and Xu}{Zhang
  et~al\mbox{.}}{2021}]%
        {zhang2021multi}
\bibfield{author}{\bibinfo{person}{Xi Zhang}, \bibinfo{person}{Feifei Zhang},
  {and} \bibinfo{person}{Changsheng Xu}.} \bibinfo{year}{2021}\natexlab{}.
\newblock \showarticletitle{Multi-Level Counterfactual Contrast for Visual
  Commonsense Reasoning}. In \bibinfo{booktitle}{\emph{Proceedings of the {ACM}
  International Conference on Multimedia}}. \bibinfo{publisher}{{ACM}},
  \bibinfo{pages}{1793--1802}.
\newblock


\bibitem[\protect\citeauthoryear{Zheng, Guo, Chen, Yu, and Jiang}{Zheng
  et~al\mbox{.}}{2020}]%
        {DBLP:conf/sigir/ZhengGCYJ20}
\bibfield{author}{\bibinfo{person}{Lin Zheng}, \bibinfo{person}{Naicheng Guo},
  \bibinfo{person}{Weihao Chen}, \bibinfo{person}{Jin Yu}, {and}
  \bibinfo{person}{Dazhi Jiang}.} \bibinfo{year}{2020}\natexlab{}.
\newblock \showarticletitle{Sentiment-guided Sequential Recommendation}. In
  \bibinfo{booktitle}{\emph{The International Conference on Research and
  Development in Information Retrieval}}. \bibinfo{publisher}{{ACM}},
  \bibinfo{pages}{1957--1960}.
\newblock


\end{thebibliography}
